\documentclass[smallextended]{svjour3}       
\smartqed  
\usepackage{graphicx}
\usepackage{hyperref}

\usepackage{placeins}
\usepackage[T1]{fontenc}
\usepackage[utf8]{inputenc}
\usepackage{amssymb}
\usepackage{amsmath}
\usepackage{enumerate}
\usepackage{bbm}
\usepackage{float}
\floatstyle{plain}
\usepackage{wrapfig}
\usepackage[usenames,dvipsnames,table,svgnames]{xcolor}
\usepackage{framed} 
\usepackage[lined, algoruled, algosection, linesnumbered, resetcount]{algorithm2e}
\usepackage{algpseudocode}
\usepackage{mdframed}
\usepackage{natbib}

\usepackage{pdflscape}

\usepackage{paralist}
\usepackage[font={small,it}, labelfont={color=RoyalBlue, bf}]{caption}

\algrenewcommand{\algorithmiccomment}[1]{$\rightarrow$ #1}
\renewcommand{\vec}[1]{\mathbf{#1}}
\newtheorem{analogy}{Intuitive Description}

\DeclareMathOperator*{\argmin}{arg\,min}

\usepackage{mathptmx} 

\journalname{Natural Computing}
\begin{document}

\title{A new Taxonomy of Global Optimization Algorithms* \thanks{*Submitted to Natural Computing, Springer}}

\author{J\"org Stork \and A. E. Eiben  \and Thomas Bartz-Beielstein}


\institute{J\"org Stork \at
	     Technische Hochschule K\"oln \\
              Steinm\"ullerallee 1, 51643 Gummersbach, Germany \\
              Tel.: +49 2261 8196-6336\\
              \email{joerg.stork@th-koeln.de}           
           \and
           A. E. Eiben \at
           Vrije Universiteit Amsterdam \\
           \email{a.e.eiben@vu.nl}  
           \and
           Thomas Bartz-Beielstein \\
           Technische Hochschule K\"oln \\
           \email{thomas.bartz-beielstein@th-koeln.de}  }

\date{Received: date / Accepted: date}
\setcounter{tocdepth}{4}

\maketitle

%
%
%

\begin{abstract} 

Surrogate-based optimization, nature-inspired metaheuristics, and hybrid combinations have become state of the art in algorithm design for solving real-world optimization problems.
Still, it is difficult for practitioners to get an overview that explains their advantages in comparison to a large number of available methods in the scope of optimization. 
Available taxonomies lack the embedding of current approaches in the larger context of this broad field.
This article presents a taxonomy of the field, which explores and matches algorithm strategies by extracting similarities and differences
in their search strategies. A particular focus lies on algorithms using surrogates, nature-inspired designs, and those created by design optimization. 
The extracted features of components or operators allow us to create a set of classification indicators to distinguish between a small number of classes. 
The features allow a deeper understanding of components of the search strategies and further indicate the close connections between the different algorithm designs.
We present intuitive analogies to explain the basic principles of the search algorithms, particularly useful for novices in this research field. 
Furthermore, this taxonomy allows recommendations for the applicability of the corresponding algorithms.

\end{abstract}




\section{Introduction}

Modern applications in industry, business, and information systems require a tremendous amount of optimization. 
Global optimization (GO) tackles various severe problems emerging from the context of complex physical systems, business processes, and particular from applications of artificial intelligence.
Challenging problems arise from industry on the application level, e.g., machines regarding manufacturing speed, part quality or energy efficiency, 
or on the business level, such as optimization of production plans, purchase, sales, and after-sales. 
Further, they emerge from areas of artificial intelligence and information engineering, such as machine learning, e.g., optimization of standard data models such as neural networks for different applications. 
Their complex nature connects all these problems: they tend to be multi-modal and expensive to solve, with unknown objective function properties, as the underlying mechanisms are often not well described or known. 

Solving such optimization problems of this kind relies necessarily on performing costly computations, simulations, or even real-world experiments, which are frequently considered being black-box. 
A fundamental challenge in such systems is the different costs of function evaluations. 
Whether we are probing a real physical system, querying the simulator, or creating a new complex data model, a significant amount of resources is needed to fulfill these tasks.
GO methods for such problems thus need to fulfill a particular set of requirements. 
They need to work with black-box style probes only, so without any further information on the structure of the problem. 
Further, they must approach the vicinity of the global optimum with a limited number of function evaluations. 

The improvement of computational power in the last decades has been influencing the development of algorithms. 
A massive amount of computational power became available for researchers worldwide through multi-core desktop machines, parallel computing, and high-performance computing clusters. 
This availability has improved the following fields of research: 
Firstly, the development of more complex, nature-inspired, and generally applicable heuristics, so-called metaheuristics. 
Secondly, it accelerated significant progress in the field of accurate, data-driven approximation models, so-called surrogate models, and their embodiment in an optimization process. 
Thirdly, the upcoming of approaches which combine several optimization algorithms and seek towards automatic combination and configuration of the optimal optimization algorithm, known as hyperheuristics. 
The hyperheuristic approach shows the close connections between different named algorithms, particular in the area of bio-inspired metaheuristics. 
Automatic algorithm composition has shown to be able to outperform available (fixed) optimization algorithms. 
All these GO algorithms differ broadly from standard approaches, define new classes of algorithms, and are not well integrated into available taxonomies.

Hence, we propose a new taxonomy, which:
\begin{enumerate}[I]
\item describes a comprehensive overview of GO algorithms, including surrogate-based, model-based and hybrid algorithms,
\item can generalize well and connects GO algorithm classes to show their similarities and differences,
\item focusses on simplicity, which enables an easy understanding of GO algorithms,
\item can be used to provide underlying knowledge and best practices to select a suitable algorithm for a new optimization problem.
\end{enumerate}

Our new taxonomy is created based on algorithm key features and divides the algorithms into a small number of intuitive classes: \emph{Hill-Climbing, Trajectory, Population, Surrogate, and Hybrid}.
Further, \emph{Exact} algorithms are shortly reviewed, but not an active part of our taxonomy, which focusses on heuristic algorithms. 
We further utilize extended class names as descriptions founded on the abstracted human behavior in pathfinding. 
The analogies \emph{Mountaineer, Sightseer, Team, Surveyor} create further understanding by using the image of a single or several persons hiking a landscape in search of the desired location (optimum) utilizing the shortest path (e.g., several iterations). 

This utilized abstraction allows us to present bright, comprehensible ideas on how the individual classes differ and moreover, how the respective algorithms perform their search. 
Although abstraction is necessary for developing our results, we will present results that are useful for practitioners. 

This article mainly addresses different kinds of readers: 
Beginners will find an intuitive and comprehensive overview of GO algorithms, especially concerning common metaheuristics and newer developments in the field of surrogate-based and hybrid and hyperheuristic optimization. 
For advanced readers, we also discuss the applicability of the algorithms to tackle specific problem properties and provide essential knowledge for reasonable algorithm selection.
We provide an extensive list of references for experienced users. 
The taxonomy can be used to create realistic comparisons and benchmarks for the different classes of algorithms. 
It further provides insights for users, who aim to develop new search strategies, operators, and algorithms. 

We organized the remainder of this article as follows: 
Section 2 presents the development of optimization algorithms and their core concepts. 
Section 3 motivates a new taxonomy by reviewing the history of available GO taxonomies, illustrates algorithm design aspects, and presents extracted classification features. 
Section 4 introduces the new intuitive classification with examples.
Section 5 introduces best practices suggestions regarding the applicability of algorithms. 
Section 6 summarizes and concludes the article with the recent trends and challenges in GO and currently essential research fields. 

\section{Modern Optimization Algorithms}
\label{sec:genalg} 
This section describes the fundamental principles of modern search algorithms, particular the elements and backgrounds of surrogate-based and hybrid optimization.

In this work, we will focus on the description of continuous search spaces, but we could transfer the presented ideas to combinatorial spaces with slight adjustments to the particularities of these problems.
The goal of global optimization is to find the overall best solution, i.e., for the common task of minimization, to discover decision variable values that minimize the objective function value. 
We denote the global search space as compact set ${\cal{S}} =\{\vec{x} \mid \, \vec{x}_l \leq \vec{x} \leq \vec{x}_u\}$ with $\vec{x}_l, \vec{x}_u \in \mathbbm{R^n}$ being the explicit, finite lower and upper bounds on $\vec{x}$.

Given an objective function f: $\mathbbm{R}^n \to \mathbbm{R}$ with real-valued input vectors $\vec{x}$ we attempt to find the location $\vec{x} \in \mathbbm{R}^n$ which minimizes the function: 
$\argmin{f(\vec{x})}, \vec{x} \in {\cal{S}}$.

Finding the global optimum is always the ultimate goal and as such desirable, but for many practical problems, a solution improving the current best solution in a given budget of evaluations or time will still be a success. 
Particular in GO the optimum commonly cannot be identified exactly; thus, modern heuristics are designed to spend their resources as efficiently as possible to approximate near-best solutions while finding the global optimum is never guaranteed. 

 \citet{Toer89a} mention three principles for the construction of an optimization algorithm:
\begin{compactenum}
\item An algorithm utilizing all available \emph{a priori} information will outperform a method using less information.
\item If no \emph{a priori} information is available, the information is completely based on evaluated candidate points and their objective values.
\item Given a fixed number of evaluated points, optimization algorithms will only differ from each other in the distribution of candidate points.
\end{compactenum}
As most modern algorithms focus on handling problems where the problem includes little or no \emph{a priori} information, the principles displayed above lead to the conclusion that the most crucial design aspect of any algorithm is to find a strategy to distribute the initial candidates in the search space and to generate new candidates based on a variation of solutions. 
These two procedures define the \emph{search strategy}, which needs to follow the two competing goals of \emph{exploration} and \emph{exploitation}. 
The balance between these two competing goals is usually part of the algorithm configuration. 
Consequently, any algorithm needs to be adapted to the structure of the problem at hand to achieve optimal performance. 
This can be considered during the construction of an algorithm, before the optimization by parameter \emph{tuning} or during the run by parameter \emph{control}~\citep{bartz2005sequential,eiben1999parameter}. 
In general, the main goal of any method is to reach their target with high efficiency, i.e., to discover optima fast and accurate with as little resources as possible.
Moreover, the goal is not mandatory finding the global optimum, which is a demanding and expensive task for many problems, but to identify a valuable local optimum or to improve the currently available solution. 
We will explicitly discuss the design of modern optimization algorithms in Section \ref{sec:algelements}.

\subsection{Exact Algorithms}
\label{sec:exact}
\emph{Exact} algorithms also referred to as \emph{non-heuristic} or \emph{complete} algorithms \citep{neumaier2004complete}, 
are a special class of \emph{deterministic, systematic} or \emph{exhaustive} optimization techniques. 
Exact algorithms have a guarantee to solve problems to optimality within using a predictable amount of resources, such as function evaluations or computation time~\citep{Fomi13a,woeginger2003exact}. 
This guarantee often requires sufficient \emph{a priori} information about the objective function. 
If they apply to a problem, these algorithms are very reliable, as they allow convergence proofs of finding the global optimum. 
Without available \emph{a priori} information, the stopping criterium needs to be defined by a heuristic approach, which softens the guarantee of solving to optimality.

\subsection{Heuristics and Metaheuristics}
In modern computer-aided optimization, heuristics and metaheuristics are well-established solution techniques. 
Although presenting solutions that are not guaranteed to be optimal, their general applicability and ability to present fast sufficient solutions make them very attractive for applied optimization. 
Their inventors built them upon the principle of \emph{trial and error}, where solution candidates are evaluated and rewarded with a \emph{fitness}. 
The term \emph{fitness} has its origins in evolutionary computation, where the fitness describes the competitive ability of an individual in the reproduction process. 
The fitness is in its purest form the objective function value $y= f(\vec{x})$ concerning the optimization goal, e.g., in a minimization problem, smaller values have higher fitness than larger values. 
Moreover, it can be part of the search strategy, e.g., scaled or adjusted by additional functions, particular for multi-objective or constrained optimization. 

\emph{Heuristics} can be defined as problem-dependent algorithms, which are developed or adapted to the particularities of a specific optimization problem or problem instance~\citep{pearl1985heuristics}. 
Typically, heuristics systematically perform evaluations, although utilizing stochastic elements. 
Heuristics use this principle to provide fast, not necessarily exact (i.e., not optimal) numerical solutions to optimization problems. 
Moreover, heuristics are often greedy to provide fast solutions but get trapped in local optima and fail to find the global optimum. 

\emph{Metaheuristics} can be defined as problem independent, general-purpose optimization algorithms. 
They apply to a wide range of problems and problem instances. 
The term \textit{meta} describes the higher-level general methodology, which is utilized to guide the underlying heuristic strategy~\citep{talbi2009metaheuristics}.

They share the following characteristics \citep{boussaid2013survey}: 
\begin{compactitem}
\item The algorithms are nature-inspired; they follow certain principles from natural phenomena or behaviors (e.g., biological evolution, physics, social behavior).
\item The search process involves stochastic parts; it utilizes probability distributions and random processes.
\item As they are meant to be generally applicable solvers, they include a set of control parameters to adjust the search strategy.
\item They do not rely on the information of the process which is available before the start of the optimization run, so-called \emph{a priori} information. Still, they can benefit from such information (e.g., to set up control parameters)
\end{compactitem}
During the remainder of this article, we will focus on heuristic, respectively, metaheuristic algorithms. 

\subsection{Surrogate-based Optimization Algorithms}
\label{sec:surrogate}
\emph{Surrogate-based optimization} algorithms are designed to process expensive and complex problems, which arise from real-world applications and sophisticated computational models. 
These problems are commonly black-box, which means that they only provide very sparse domain knowledge. 
Consequently, problem information needs to be exploited by experiments or function evaluations. 

Surrogate-based optimization is developed to exhaust the available information by utilizing a \emph{surrogate model} optimally. 
A surrogate model is an approximation which substitutes the original expensive objective function, real-world process, physical simulation, or computational process during the optimization.  
In general, surrogates are either simplified \emph{physical} or numerical models based on knowledge about the physical system, or empirical \emph{functional} models based on knowledge acquired from evaluations and sparse sampling of the parameter space \citep{sondergaard2003optimization}. 
In this work, we focus on the latter, so-called data-driven models.
The terms \emph{surrogate model}, \emph{meta-model}, \emph{response surface model} and also \emph{posterior distribution} are used synonymously in the common literature~\citep{mockus1974bayesian,jones2001taxonomy,bartz2017model}. 
We will briefly refer to a surrogate model as a \emph{surrogate}. 
Furthermore, we assume that it is crucial to distinguish between the use of an explicit surrogate of the objective function and general \emph{model-based} optimization \citep{zlochin2004model}, which additionally refers to methods, where a statistical model is used to generate new candidate solutions (cf. Section \ref{sec:algelements}). 
We thus distinguish between the two different terms \emph{surrogate-based} and \emph{model-based} to avoid confusion. 
Another term present in the literature is \emph{surrogate-assisted} optimization, which mostly refers to the application of surrogates in combination with population-based evolutionary computation~\citep{jin2011surrogate}. 

\begin{figure}[ht]
\sidecaption
\centering
\includegraphics[width= 0.7\columnwidth] {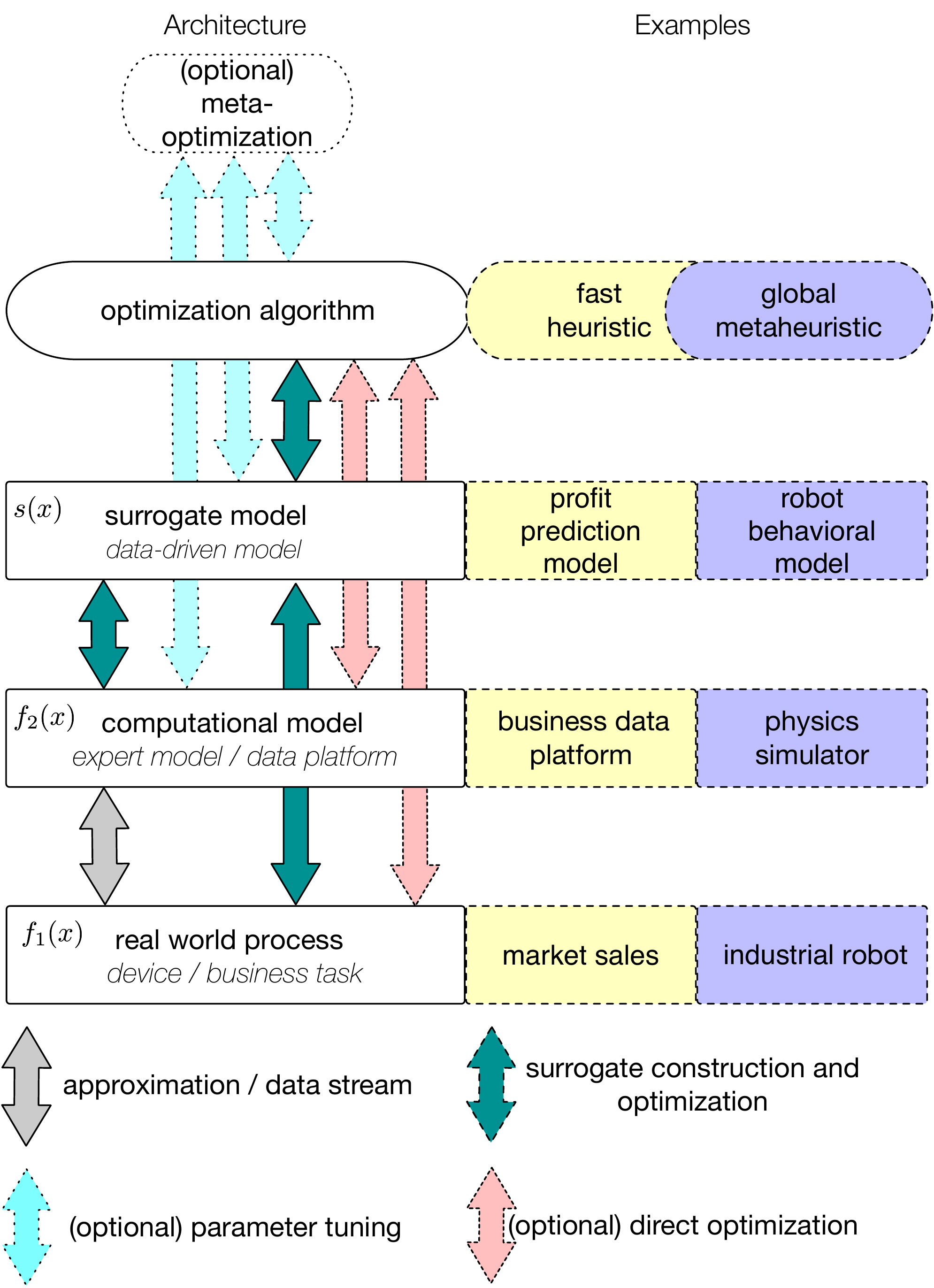}
\caption{A surrogate-based optimization process with the different objective function layers: real-world process, computational model and surrogate.
The arrows mark different possible data/application streams. Two examples of processes are given to outline the use of the architecture in a business data and robot control task.
}
\label{fig:classSurrogate}
\end{figure}

Important publications featuring overviews or surveys on surrogates and surrogate-based optimization were presented by \citet{sacks1989design}, \citet{jones2001taxonomy}, \citet{queipo2005surrogate}, \citet{forrester2009recent}.
Surrogate-based optimization is commonly defined for but not limited to the case of complex real-world optimization applications, where we identify two typical problem layers and a surrogate layer. 
We could transfer the defined layers to different computational problems with expensive function evaluations, such as complex algorithms or machine learning tasks.
Each layer can be the target of optimization or used to retrieve information to guide the optimization process. 
Figure \ref{fig:classSurrogate} illustrates the different layers of objective functions and the surrogate-based optimization process for real-world problems. In this case, the objective function layers, from the bottom up, are:
\begin{enumerate}[L1] 
\item The real-world application $f_1(\vec{x})$, given by the physical process itself or a physical model. Direct optimization is often expensive or even impossible, due to evaluations involving resource-demanding prototype building or even dangerous experiments. 
\item The computational model $f_2(\vec{x})$, given by a simulation of the physical process or a complex computational model, e.g., a computational fluid dynamics model or structural dynamics model. A single computation may take minutes, hours, or even weeks to compute. 
\item The surrogate $s(\vec{x})$, given by a data-driven regression model. The accuracy heavily depends on the underlying surrogate type and number of available information (i.e., function evaluations). The optimization is, compared to the other layers, typically cheap. Surrogates are constructed either for the real-world application $f_1(\vec{x})$ or the computational model $f_2(\vec{x})$.
\end{enumerate}
Furthermore, the surrogate-based optimization cycle includes the optimization process itself, which is given by an adequate optimization algorithm for the selected objective function layer. 
No surrogate-based optimization is performed, if the optimization is directly applied to $f_1(\vec{x})$ or $f_2(\vec{x})$. 
The surrogate-based optimization uses $f_1(\vec{x})$ or $f_2(\vec{x})$ for verification of promising solution candidates. 
Moreover, the control parameters of the optimization algorithm or even the complete optimization cycle, including the surrogate modeling process, can be tuned.  

Each layer imposes different evaluation costs and fidelities: the real-world problem is the most expensive to evaluate, but has the highest fidelity, while the surrogate is the cheapest to evaluate, but has a lower fidelity. 
The main benefit of using surrogates is thus the reduction of needed expensive function evaluations on the objective function $f_1(\vec{x})$ or $f_2(\vec{x})$ during the optimization. 
The studies by \citet{loshchilov2012self}, \citet{marsden2004optimal}, \citet{ong2005surrogate} and \citet{won2004performance} feature benchmark comparisons of surrogate-based optimization.
Nevertheless, the model construction and updating of the surrogates also require computational resources, as well as evaluations for verification on the more expensive function layers. 
An advantage of surrogate-based optimization is the availability of the surrogate model, which can be utilized to gain further global insight into the problem, which is particularly valuable for black-box problems. 
For instance, the surrogate can be utilized to identify important decision variables or visualize the nature of the problem, i.e., the fitness landscape. 

\subsection{Meta-Optimization and Hyperheuristics}
\emph{Meta-optimization} or \emph{parameter tuning}~\citep{mercer1978adaptive} describes the process of finding the optimal parameter set for an optimization algorithm.
It is also an optimization process itself, which can become very costly in terms of objective function evaluations, as they are required to evaluate
the parameter set of a specific algorithm.
Hence, particular surrogate-model based algorithms have become very successful meta-optimizer~\citep{bartz2005sequential}.
Figure \ref{fig:classSurrogate} shows where the meta-optimization is situated in an optimization process. 
If the algorithm adapts parameters during the active run of optimization, it is called parameter control \citep{eiben1999parameter}. 
Algorithm parameter tuning and control is further discussed in Section {\ref{sec:algelements}}.

A \emph{hyperheuristic} \citep{cowling2000hyperheuristic,cowling2002hyperheuristics} is a high-level approach that selects and combines low-level approaches (i.e., heuristics, elements from metaheuristics), to solve a specific problem or problem class. 
It is an optimization algorithm that automates the algorithm design process by searching an ample space of pre-defined algorithm components.
A hyperheuristic can also be utilized in an online fashion, e.g., trying to find the most suitable algorithm at each state of a search process \citep{Vermetten19a}.
We regard hyperheuristics as hybrid algorithms (cf. Section \ref{sec:hybrid}).
 
\section{A New Taxonomy}
\label{sec:class}

The term \emph{taxonomy} is defined as a consistent procedure or classification scheme for separating objects into classes or categories based on specific features. 
The term taxonomy is mainly present in natural science for establishing hierarchical classifications. 
A taxonomy fulfills the task of distinction and order; it provides explanations and a greater understanding of the research area through the identification of coherence and the differences between the classes. 

Several reasons drive our motivation for a new taxonomy:
The first reason I) is that considering available GO taxonomies (Section \ref{sec:hist}, cf. Figure \ref{fig:classHist}), we can conclude that during the last decades, several authors developed new taxonomies for optimization algorithms. 
However, new classes of algorithms have become state-of-the-art in modern algorithm design, particularly model-based, surrogate-based, and hybrid algorithms dominate the field. 
Existing taxonomies of GO algorithms do not reflect this situation. 
Although there are surveys and books which handle the broad field of optimization and give general taxonomies, they are outdated and lack the integration of the new designs.
Available up-to-date taxonomies often address a particular subfield of algorithms and discuss them in detail.
However, a generalized taxonomy, which includes the above-described approaches and allows to connect these optimization strategies, is missing. 

This gap motivated our second reason II) the development of a generalization scheme for algorithms.
We argue that the search concepts of most algorithms are similar.
While the algorithms have apparent differences in their strategies, they are not overall different algorithms.
Certain elements are characteristic of algorithms, which allow us to define classes based on their search elements.
Even different classes share a large amount of these search elements.
Thus our new taxonomy is based on a generalized scheme of five crucial algorithm design elements (Section \ref{sec:algelements}, cf. Figure \ref{fig:classFeatures}), which allows us to take a bottom to top approach to differentiate, but also connect the different algorithm classes. 
The recent developments in hybrid algorithms drive the urge to generalize search strategies, 
where we no longer use specific, individual algorithms, but combinations of search elements and operators of different classes to find and establish new strategies, which cannot merely be categorized. 

Our third reason III) is the importance of simplicity. 
Our new taxonomy is not only intended to divide the algorithms into classes, but also to provide an intuitive understanding of the working mechanisms of each algorithm to a broad audience.
To support these ideas, we will draw \emph{analogies} between the algorithm classes and the behavior of a human-like individual in each of the descriptive class sections. 

Our last reason IV) is that we intend our taxonomy to be helpful in practice. 
A common issue is the selection of an adequate optimization algorithm if faced with a new problem. 
Our algorithm classes are connected by individual properties, 
which allows us to utilize the new taxonomy to propose suitable algorithm classes based on a small set of problem features. 
These suggestions, in detail discussed in section \ref{sec:algselect} and illustrated in figure \ref{fig:classSelection} shall help users to find best practices for new problems.

\subsection{History of Taxonomies} 
\label{sec:hist}
\begin{figure}
\includegraphics[width= 1\textwidth] {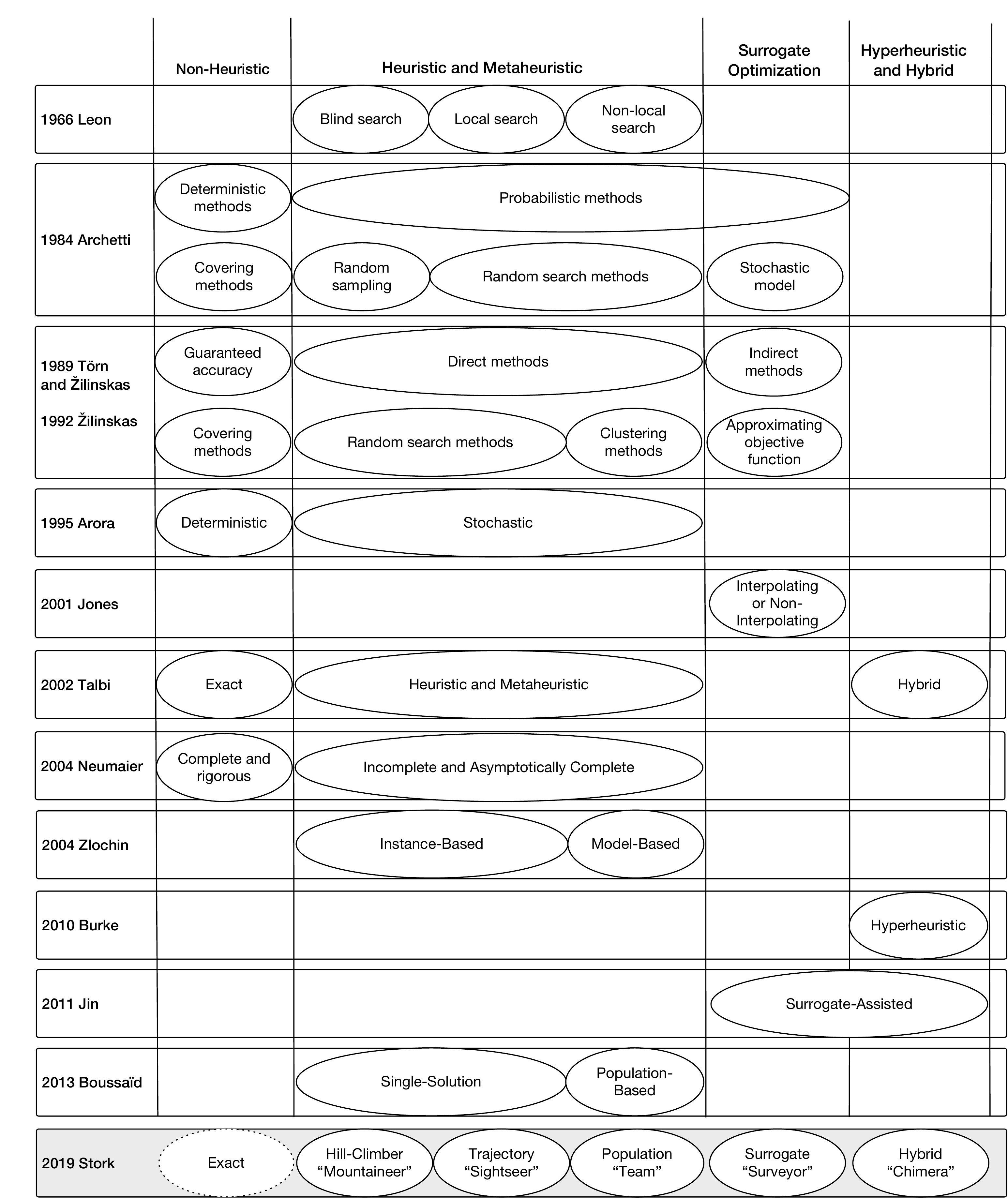}
\caption{Global Optimization Taxonomy History. Information from \citet{leon1966classified}, \citet{Arch84a}, \citet{Toer89a},\citet{arora1995global}, \citet{jones2001taxonomy}, \citet{talbi2002taxonomy}, \citet{neumaier2004complete}, \citet{zlochin2004model},\citet{burke2010classification}, \citet{jin2011surrogate} and \citet{boussaid2013survey} are illustrated and compared. Different distinctions between the large set of algorithms were drawn. A comprehensive taxonomy is missing and introduced by our new taxonomy, which concludes the diagram and is further presented in Section \ref{sec:class}.}
\label{fig:classHist}
\end{figure}

In the literature, one can find several taxonomies trying to shed light on the vast field of optimization algorithms.
The identified classes are often named by a significant feature of the algorithms in the class, with the names either being informative or descriptive.
For example, \citet{leon1966classified} presented one of the first overviews on global optimization. 
It classified algorithms into three categories: 1. \emph{Blind search}, 2. \emph{Local search} 3. \emph{Non-local search}.
In this context, \emph{blind search} refers to simple search strategies that select the candidates at random over the entire search space, but following a built-in sequential selection strategy. 
During the \emph{local search}, new candidates are selected only in the immediate neighborhood of the previous candidates, which leads to a trajectory of small steps. 
Finally, \emph{non-local search} allows to escape from local optima and thus enables a global search. 
\citet{Arch84a} extends the above scheme by also adding the class of \emph{deterministic} algorithms, i.e., those who are guaranteed to find the global optimum with a defined budget.
Furthermore, the paper stands out in establishing a taxonomy, which for the first time includes the concepts to construct surrogates, as they describe \emph{probabilistic} methods based on statistical models, which are iteratively utilized to perform the search. 
\citet{Toer89a} reviewed existing classification schemes and presented their classifications. 
They made that the most crucial distinction between two non-overlapping main classes, namely those methods with or without guaranteed accuracy. 
The main new feature of their taxonomy is the clear separation of the heuristic methods in those with \emph{direct} and \emph{indirect} objective function evaluation.
\citet{mockus1974bayesian} also discussed the use of Bayesian optimization. 
Today's high availability of computational power did not exist; therefore, \citet{Toer89a} concluded the following regarding Bayesian models and their applicability for (surrogate-based) optimization: 
\begin{quote}
Even if it is very attractive, theoretically it is too complicated for algorithmic realization. Because of the fairly cumbersome computations involving operations with the inverse of the covariance matrix and complicated auxiliary optimization problems the resort has been to use simplified models.
\end{quote}
Still, we find the scheme of dividing algorithms into non-heuristic (or exact), random (or stochastic) and further surrogate-based frequently. 
Several following taxonomies added different algorithm features to their taxonomies, such as metaheuristic approaches \citep{arora1995global}, surrogate-based optimization \citet{jones2001taxonomy}, non-heuristic methods \citet{neumaier2004complete}, hybrid methods \citet{talbi2002taxonomy}, direct search methods \citep{audet2014survey,kolda2003optimization}, model-based optimization \citet{zlochin2004model}, hyperheuristics \citet{burke2010classification}, surrogate-assisted algorithms \citet{jin2011surrogate}, nature-inspired methods \citep{rozenberg2011handbook}, or population-based approaches \citet{boussaid2013survey}.
We created an overview of different selected taxonomies and put them into the comparison in Figure \ref{fig:classHist}.

\subsection{The Four Elements of Algorithm Design}
\label{sec:algelements}

Any modern optimization algorithm, as defined in Section \ref{sec:genalg}, can be reduced to the three key search strategy elements \emph{Initialization}, \emph{Generation} and \emph{Selection}. 
A fifth element controls all these key elements: the \emph{Control} of the different functions and operators in each element.
The underlying terminology is generic and based on typical concepts from the field of evolutionary algorithms. 
We could easily exchange it with wording from other standard algorithm classes (e.g., evaluate= test/trial, generate=produce/variate).
Algorithm \ref{alg:optim} displays the principal elements and the abstracted fundamental structure of optimization algorithms \citep{bartz2017model}. 
We could map this structure and elements to any modern optimization algorithm. 
Even if the search strategy is inherently different or elements do not follow the illustrated order or appear multiple times per iteration. 

The \textbf{initialization} of the search defines starting locations or a schema for the initial candidate solutions. 
Two common strategies exist: 
\begin{enumerate}
\item If there is no a priori knowledge about the problem and its search space, the best option is to use strategically randomized starting points. Particularly interesting for surrogate-based optimization are systematic initialization schemes by methods from the field of \emph{design of experiments}.
\item If domain knowledge or other a priori information is available, such as information from the data or process from previous optimization runs, the algorithm should be initialized utilizing this information to a full extent, e.g., by using a selection of these solutions, such as these with the best fitness. In surrogate-based optimization, the initial modeling can use available data. 
\end{enumerate}
The initial candidates have a large impact on the balance between exploration and exploitation. 
Space-filling designs with large amounts of random candidates or sophisticated \emph{design of experiments} methods will lead to an initial exploration of the search space. 
Starting with a single candidate will presumably lead to an exploitation of the neighborhood of the selected candidate location. 
Hence, algorithms using several candidates are in general more robust, while a single candidate algorithms are sensitive to the selection of the starting candidate, particular in multi-modal landscapes. 
Multi-start strategies can further increase the robustness and are particularly common for single-candidate algorithms, and also frequently recommended for population-based algorithms~\citep{hansen2010comparing}.

\begin{algorithm}
 \caption{General Optimization Algorithm}
 \label{alg:optim}
 \textbf{set} initial \emph{control parameters} \\
 \Begin{
 $t=0$ \\
 \textbf{initialize} candidate(s) \\
 \emph{evaluate} initial candidate(s)\\
 \While{\textbf{not} termination-condition}{
 $t= t+1$ \\
 \textbf{generate} new candidate(s) \\
 \emph{evaluate} new candidate(s) \\
 \textbf{select} solution(s) for next iteration \\
 \emph{\textbf{optional:} update control parameters} \\
  }}
\end{algorithm}

The \textbf{generation} during the search process defines the methods for finding new candidates, with particular regard on how they use available or obtained information about the objective function. 
A standard approach is the variation of existing observations, as it utilizes, and to a certain extent preserves, the information of previous iterations. 
Even by the simplest \emph{hill-climber} class algorithms, which do not require any global information or stored knowledge of former iterations (Sec \ref{sec:hill-climber}), use the last obtained solution to generate new candidate(s).  
Sophisticated algorithms generate new candidates based on exploited and stored global knowledge about the objective function and fitness landscape. 
This knowledge is stored by either keeping an archive of all available or selected observations or implicitly by using distribution or data models of available observations. 
Another option to generate new candidates is combining information of multiple candidates by dedicated functions or operators, particular present in the trajectory class (Sec \ref{sec:trajectory}). 
The exact operators for generation and variation of candidate solutions are various and an essential aspect of keeping the balance between exploration and exploitation in a search strategy.

The \textbf{selection} defines the principle of choosing the solutions for the next iteration. 
We use the term \emph{selection}, which has its origins in evolutionary computation. 
Besides the most straightforward strategy of choosing the solution(s) with the \emph{best} fitness, advanced selection strategies have emerged, which are mainly present in metaheuristics~\citep{boussaid2013survey}. 
These selection strategies are particularly common in algorithms with several candidates per \emph{generation} step; thus, evolutionary computation introduced the most sophisticated selection methods~\citep{eiben2015introduction}. 
The use of absolute differences in fitness or their relative difference is the most common strategy and called \emph{ranked} selection. 

\textbf{Control parameters} determine how the search can be adapted and improved by controlling the above mentioned key elements. 
We distinguish between \emph{internal} and \emph{external} parameters: External parameters, also known as offline parameters, can be adjusted by the user and need to be set a priori to the optimization run. 
Typical external parameters include the number of candidates and settings influencing the above mentioned key elements. 
Besides standard theory-based defaults~\citep{schwefel1993evolution}, they are usually set by either utilizing available domain knowledge, extensive a priori benchmark experiments~\citep{gamperle2002parameter}, or educated guessing. 
Sophisticated \emph{meta-optimization} methods were developed to exploit the right parameter settings in an automated fashion. 
Well-known examples are sequential parameter tuning~\citep{bartz2005sequential}, iterated racing for automatic algorithm tuning~\citep{lopez2016irace}, \emph{bonesa}~\citep{smit2011multi} or \emph{SMAC}~\citep{hutter2011sequential}. 
In comparison to external parameters, internal ones are not meant to be changed by the user. 
They are either fixed to an absolute value, which is usually based on physical constants or extensive testing by the authors of the algorithm, or are \emph{self-adaptive}. 
Self-adaption or online control changes the parameters online during the search process based on the gathered knowledge or exploited problem information~\citep{eiben1999parameter} without user influence. 
Algorithms using self-adaptive schemes thus tend to gain outstanding generalization abilities and are especially interesting for black-box problems, where no information about the objective function properties is available~\citep{hansen2003reducing}. 
In general, the settings of algorithm control parameters directly affect the balance between exploration and exploitation during the search and are crucial for the performance.

Further, the \emph{evaluation} step computes the fitness of the candidates. 
Of course, the evaluation is a crucial aspect, as it defines the basis for any information and also influences the search strategy. 
However, we do not regard the evaluation as a central part of the algorithm design, as it is mostly problem-dependent. 
Important aspects of evaluation are \emph{noise}, \emph{constraints} and \emph{multiple objectives}. 
While most computer experiments are deterministic, i.e., iterations using the same value set for the associated decision variables should deliver the same results, real-world problems are often non-deterministic. 
They include non-observable disturbance variables and stochastic \emph{noise}. 
Typical noise handling techniques include multiple evaluations of solutions to reduce the standard deviation and special sampling techniques. 
The interested reader can find a survey on noise handling by \citet{arnold2003comparison}.
Moreover, many real-world problems frequently include different constraints, which we need to consider during the optimization process. 
Constraint handling techniques can be directly part of the optimization algorithm, but most algorithms are designed to minimize the objective function and add constraint handling on top. 
Thus, algorithms integrate it by adjusting the fitness, e.g., by penalty terms. 
Different techniques for constraint handling are discussed by \citet{coello2002theoretical} and \citet{arnold20121}.
The evaluation of multiple objectives can include several correlated objective functions and usually delivers a set of non-dominated solutions, a so-called Pareto-set~\citep{naujoks2005multi}. 
In this case, a so-called decision-maker computes the fitness of the solutions and selects reasonable solutions from the Pareto-set~\citep{fonseca1993genetic}. 

\section{The Definition of Intuitive Algorithms Classes}
In his work about evolution strategies, \citet{rechenberg1994} illustrated a visual approach to an optimization process: a mountaineer in an alpine landscape, attempting to find and climb the highest mountain. 
The usage of analogies to the natural world is a valuable method to explain the behavior of search algorithms.
In the area of metaheuristics, the behavior of the nature and animals inspired the search procedure of the algorithms: Evolutionary algorithms follow the evolution theory \citep{rechenberg1994,eiben2015introduction}; particle swarm optimization \citep{kennedy1995particle,shi1998modified} utilizes a strategy similar to the movement of bird flocks; ant colony optimization \citep{dorigo2006ant} mimics, as the name suggests, the ingenious pathfinding and food search principles of ant populations. 

We take up the idea of optimization processes being human-like individuals and use it in the definition of our extended class names: the \emph{mountaineer}, \emph{sightseer}, \emph{team}, \emph{surveyor} and \emph{chimera}. 
This additional naming shall accomplish the goal of presenting an evident and straightforward idea of the search strategies of the algorithms in the associated class.

\subsection{Hill-Climbing Class: "The Mountaineer"}
\label{sec:hill-climber}
\begin{analogy}[The Mountaineer] \ \\
The mountaineer is a single individual who hikes through a landscape, with a concentration on achieving his ultimate goal: finding and climbing the highest mountain. 
He is utterly focussed on his goal to climb up that mountain so that he does not consider to leave an ascending way or to explore the remaining landscape.
\end{analogy}
\emph{Hill-Climbing} algorithms focus their search strategy on greedy exploitation with minimal exploration.
Hence, this class encompasses fundamental, basic optimization algorithms with direct search strategies, which include classical gradient-based algorithms as well as deterministic or stochastic hill-climbing algorithms.
These algorithms have, by design, fast convergence to a local optimum situated in a region of attraction and commonly no explicit strategy for exploration. 
Overviews of associated algorithms were presented by \citet{lewis2000direct} and \citet{kolda2003optimization}. 
Common algorithms include the quasi-Newton \emph{Broyden-Fletcher–Goldfarb-Shanno} algorithm \citep{shanno1970conditioning}, \emph{conjugate gradients} (CG) \citep{fletcher1976conjugate},
the direct search algorithm \emph{Nelder-Mead} \citep{nelder1965simplex}, and stochastic hill climbers such as the \emph{(1+1)-Evolution Strategy} \citep{rechenberg1973evolutionsstrategie,schwefel1977numerische}.
As this class defines fundamental search strategies, hill-climbers are often part of sophisticated algorithms as a fast-converging local optimizer. 

Hill-climbers do not utilize individual operators for the initialization of the single starting point.
Thus, it is typically selected at random in the valid search space or based on prior knowledge.

The variation of the last observed selected candidate generates new candidates, commonly in the vicinity of the current solution.
For example, the stochastic hill climber utilizes random variation with small step size, in comparison to the range of ROI.
Gradient-based methods directly compute or approximate the gradients of the objective function to find the best direction and strength for the variation. 
Algorithms such as Nelder-Mead create new candidates by computing a search direction using simplexes. 

The most common selection methods are elitist strategies, which evaluate the new candidate, compare it to the old solution, and keep the one with the best fitness as a new solution. 
Always selecting the best is known as a greedy search strategy, as it tries to improve as fast as possible.
This greedy strategy leads to the outlined \emph{hill-climbing} search which performs a trajectory of small, fitness-improving steps, which forms in the ideal case a direct line to the nearest optimum. 
In general, these algorithms search locally for an optimum and do not exploit or use global function information.

Control parameter choices less influence the utilized generation and selection methods if they even require parameters.
The most critical parameter is the variation step size, which directly influences the speed of convergence.
As a result of this, the state of the art is to use an adaptive variation step size that changes online during the search, 
often based on previous successful steps, for example as defined in the famous \emph{1/5 rule}~\citep{rechenberg1973evolutionsstrategie}. 

\subsection{Trajectory Class: "The Sightseer"} 
\label{sec:ssbg}
\label{sec:trajectory}
\begin{analogy}[The Sightseer] \ \\
The intuitive idea of this class is a single or group of hiker(s) looking for interesting places. 
During their search, the sightseer takes into account that multiple places of interest exist.
They thus explore the search space or systematically visit areas to gather information about multiple locations and utilize this to find the most desired ones. 
\end{analogy}
\emph{Trajectory} class algorithms still focus on exploitation but are supported by defined exploration methods.
This class encompasses algorithms that utilize information from consecutive function evaluations. 

They are the connecting link between the hill-climbing and population class. 
While trajectory algorithms are a step towards population algorithms and also allow the sampling of several solutions in one iteration, they use the principle of initializing and maintaining a single solution.
This solution is the basis for variation in each iteration.
Again, this variation forms a trajectory in the search space over consecutive iterations, similar to the hill-climbing class.
Thus these methods are known as \emph{trajectory methods} \citep{boussaid2013survey}. 
While the initialization and generation of the trajectory class are similar to those of the hill-climbing class, the 
main differences can be found during the \emph{selection}, as they utilize operators to guide the search process in a global landscape in specific directions.
Two different strategies can be differentiated, which define two subclasses:
\begin{enumerate}[(i)]
\item The \emph{exploring trajectory} class utilizes parameter-driven acceptance functions 
\item The \emph{systematic trajectory} class utilizes a separation of the search space 
\end{enumerate}
These algorithms are susceptible to the correct parametrization, which need to be selected adequate to the underlying objective function.

\subsubsection{Exploring Trajectory Algorithms}
The exploring trajectory subclass encompasses algorithms that implement \emph{selection} operators to balance exploration and exploitation to enable global optimization.
The introduction of selection functions that allow to expand the search space and escape the \emph{region of attraction} of a local optimum achieves exploration. 
\emph{Simulated annealing}\/ (SANN) \citet{kirkpatrick1983optimization}, which is known to be a fundamental contribution to the field of metaheuristic search algorithms, exemplifies this class.
The continuous version~\citep{goffe1994global,siarry1997enhanced,van1998constrained} of the SANN algorithm extends the iterated stochastic hill-climber. 
It includes a new element for the \emph{selection}, the so-called acceptance function.
It determines the probability of accepting an inferior candidate as a solution by utilizing a parameter called temperature, in analogy to metal annealing procedures. 
This dynamic \emph{selection} allows escaping local optima steps by accepting movement in the opposite direction of improvement, which is the fundamental difference to a hill-climber and ultimately allows the global search.
At the end of each iteration, a so-called \emph{cooling} operator adapts the temperature.
This operator can be used to further balance the amount of exploration and exploitation during the search. \citep{henderson2003theory}. 
A common approach is to start with a high temperature and steadily reduce T according to the number of iterations or to utilize an exponential decay of T. 
This steady reduction of T leads to a phase of active movement and thus exploring in the early iterations, while with decreasing T, the probability of accepting inferior candidates reduces. 
With approaching a T value of zero, the behavior becomes similar to an iterative hill-climber. 
Modern SANN implementations integrate \emph{self-adaptive} cooling-schemes which use alternating phases of cooling and reheating \citep{locatelli2002simulated}. 
These allow alternating phases of exploration and exploitation but require sophisticated control. 

\subsubsection{Systematic Trajectory Algorithms}

This subclass encompasses algorithms, which base their search on a space partitioning utilizing the exposed knowledge of former iterations. 
They create sub-spaces that are excluded from \emph{generation} and \emph{selection}, or \emph{attractive} sub-spaces, where the search is focused on. 
These search space partitions guide the search by pushing candidate generation to new promising or previous unexplored parts of the search space. 
An outstanding paradigm for this class is \emph{Tabu Search} \citep{glover1989tabu}.
A so-called tabu list contains the last successful candidates and defines a sub-space of all evaluated solutions. 
In the continuous version, \citet{siarry1997fitting,hu1992tabu,chelouah2000tabu}, small (hypersphere or hyperrectangle) regions around the candidates are utilized.
The algorithm will consider these solutions or areas as forbidden for future searches, i.e., it selects no candidates situated in these regions as solutions.
This process shall ensure to move away from known solutions and prevents identical cycling of candidates and getting stuck in local optima. 
 The definition of the tabu list parameters can control exploration and exploitation by, e.g., by the number of elements or size of areas. 

The areas of search can also be pre-defined, such as in \emph{variable neighborhood search (VNS)} \citep{hansen2003variable,hansen2010variable,mladenovic2008general}.
The search strategy of VNS is to perform sequential local searches in these sub-spaces to exploit their local optima. 
The idea behind this search strategy is that by using an adequate set of sub-spaces, the chance of exploiting a local optimum, which is near the global optimum, increases.

\subsection{Population Class: "The Team"}
\begin{analogy}[The Team] \ \\
The intuitive idea of this class is a group of individuals, which \emph{team} up to achieve their mutual goal together.
They split up to explore different locations and share their knowledge with other members of the team. 
\end{analogy}
\emph{Population} class algorithms utilize distributed exploration and exploitation.
The idea of initializing, variation, and selection of several contemporary candidate solutions defines this class.
The algorithms are commonly metaheuristics, whose search concepts follow processes found in nature. 
Moreover, it includes algorithms building upon the population-based concept by utilizing models of the underlying candidate distributions. 
Due to utilizing a population, the \emph{generation} and \emph{selection} strategies of these algorithms differ significantly from the hill-climber und trajectory class.  
We subdivide this class into the classic population and model-based population algorithms, which
particularly differ in how they generate new candidates during the search:
\begin{enumerate}[(i)]
\item The \emph{classic population}\/ (Section~\ref{sec:pop}) generate and maintain several candidates with specific population-based operators.
\item The \emph{model-based population}\/ (Section~\ref{sec:mbg}) generate and adapt distribution models to store and process information. 
\end{enumerate}

\subsubsection{Classic Population Algorithms}
\label{sec:pop}
Well-known examples of this class are \emph{particle swarm optimization (PSO)} \citep{kennedy1995particle,shi1998modified} and different \emph{evolutionary algorithms (EA)}. 
We regard EAs as state of the art in population-based optimization, as their search concepts are dominating for this field. 
Nearly all other population-based algorithms use similar concepts and are frequently associated with EAs. \citet{fleming2002evolutionary} go as far to state: 
\begin{quote}
In general, any iterative, population-based approach that uses selection and random variation to generate new solutions can be regarded as an EA. 
\end{quote}
Evolutionary algorithms follow the idea of evolution, reproduction, and the natural selection concept of \emph{survival of the fittest}.
In general, the field of EAs goes back to four distinct developments, \emph{evolution strategies} (ES) \citep{rechenberg1973evolutionsstrategie,schwefel1977numerische}, \emph{evolutionary programming} \citep{fogel1966artificial}, \emph{genetic algorithms} \citep{holland1992adaptation}, and \emph{genetic programming} \citep{koza1992genetic}.
The naming of the methods and operators matches with their counterparts from biology: candidates are \emph{individuals} who can be \emph{selected} to take the role of \emph{parents}, mate and \emph{recombine} to give birth to \emph{offspring}. 
The population of individuals is \emph{evolved} (varied, evaluated, and selected) over several iterations, so-called generations, to improve the solutions. 

Different overview articles shed light on the vast field of evolutionary algorithms \citep{back1997handbook,eiben2003introduction,eiben2015introduction}. 

EAs \emph{generate} new solutions typically by variation of a selected subset of the entire population. Typically, competition-based strategies, which also often includes probabilistic elements, select the subsets.
Either random variation of this subpopulation or \emph{recombination} by crossover, which is the outstanding concept of EAs, generates new candidates. 
Recombination partly swaps the variables of two or more candidates, aggregated or combined to create new candidate solutions.

The population-based \emph{selection} strategies allow picking solutions with inferior fitness for the variation process, which allows exploration of the search space. 
Several selection strategies exist.
 For instance, in \emph{roulette} wheel selection, the chance of being selected is proportional to the ranking while all chances sum up to one. 
A spin of the roulette wheel chooses each candidate, where the individual with the highest fitness also has the highest chance of being selected. 
Alternatively, in \emph{tournament selection}, different small subsets of the population are randomly drawn for several tournaments. 
Within these tournaments, the candidates with the best fitness are selected based on comparisons to their competitors.
This competition-based selection also allows inferior candidates to win their small tournament and participate in the variation. 

EAs usually have several parameters, such as the selection pressure (i.e., the proportion of selected candidates), variation step size, or recombination probability. 
Parameter settings, particular adaptive and self-adaptive control for evolutionary algorithms is discussed in \citet{angeline1995adaptive,eiben1999parameter,lobo2007parameter}. 
 
\subsubsection{Model-based Population Algorithms}
\label{sec:mbg}
The model-based population class encompasses algorithms, which explicitly use mathematical or statistical models of the underlying candidate distributions. 
While these algorithms generally belong to the broad field of EAs (Section \ref{sec:pop}), and use similar terminology and also operators,
they are known as \emph{estimation of distribution algorithms} (EDA) \citep{larranaga2001estimation}. 
Common examples of model-based algorithms are the \emph{covariance matrix adaption - evolution strategy} (CMA-ES) \citep{hansen2003reducing} and \emph{ant colony optimization for continuous domains} $ACO_{\mathbb{R}}$ \citep{dorigo2006ant,socha2008ant}.
The algorithms can utilize different models, e.g., Bayesian networks or multivariate Gaussian distributions.  
The general idea behind the usage of a model is that it is beneficial to learn the structural information of the underlying population~\citep{larranaga2001estimation,hauschild2011introduction}.  
The distinction to other model-building classes, such as surrogate algorithms, is that the distribution models do not depict global approximations of the underlying objective function.
Their target of these models is not to replace the entire fitness landscape. 

Instead, they are being used mainly for the \emph{generation} of new candidates, which is also the main difference to non-model-based EAs.
Evolutionary operators, such as recombination or mutation, are not directly applied to the population, but a distribution model of the population spread.
For example, typical parameters of a multivariate Gaussian distribution are the mean, variance, and covariance of the population.
The algorithm uses the distribution to sample new candidates, often after the model parameters were the target of variation. 
Thus, the distribution of the candidates is changed, but not individual candidates as in an EA.

This class again has several \emph{control} parameters, which are also often designed to be adaptive or self-adaptive.
For example, the CMA-ES adapts the parameters during each iteration following the history of prior successful iterations, the so-called \emph{evolution paths}. 
These evolution paths are exponentially smoothed sums for each distribution parameter over the consecutive prior iterative steps. 
They thus utilize the information of several successful search steps. 

\subsection{Surrogate Class: "The Surveyor"}
\label{sec:surrogateclass}
\label{sec:mapper}

\begin{analogy}[The Surveyor] \ \\
The intuitive idea of the surveyor is a specialist who systematically measures a landscape by taking samples of the height to create a topological map. 
This map resembles the real landscape with a given approximation accuracy and is typically exact at the sampled locations and models the remaining landscape by regression. 
It can then be examined and utilized to approximate the quality of an unknown point and further be updated if new information is acquired.
Ultimately it can be used to guide an individual to the desired location.
\end{analogy}
\emph{Surrogate} class algorithms utilize distributed exploration and exploitation by explicitly relying on landscape information and a landscape model.
These algorithms differ from all other defined classes in their focus on acquiring, gathering, and utilizing information about the fitness landscape. 
They utilize evaluated, acquired information to approximate the landscape and also predict the fitness of new candidates. 

As illustrated in Section \ref{sec:surrogate}, the surrogates depict the \emph{maps} of the fitness landscape of an objective function in an algorithmic framework. 
A surrogate algorithm utilizes them for an efficient indirect search, instead of performing multiple, direct, or localized search steps. 
We divide this class into two subclasses:
\begin{itemize}
\item \emph{Surrogate-based} algorithms utilize a global surrogate model for variation and selection.
\item \emph{Surrogate-assisted} algorithms utilize surrogates to improve selection in population-based algorithms.
\end{itemize}

\label{sec:models}
For both classes, the surrogate model is a core element of any variation and selection process during optimization and essential for their performance. 
A perfect surrogate provides an excellent fit to observations, while ideally possessing superior interpolation and extrapolation abilities. 
However, a large number of available surrogate models all have significant differing characteristics, advantages, and disadvantages. 
Model selection is thus a complicated and challenging task. 
If no domain knowledge is available, such as in real black-box optimization, it is often inevitable to test different surrogates for their applicability. 

Common models are: linear, quadratic or polynomial regression, Gaussian processes (also known as Kriging) \citep{sacks1989design,forrester2008engineering}, regression trees \citep{breiman1984classification}, artificial neural networks and radial basis function networks \citep{haykin2004comprehensive,hornik1989multilayer} including deep learning networks \citep{collobert2008unified,hinton2012deep,hinton2006fast} and symbolic regression models \citep{augusto2000symbolic,2010flasch,mckay1995using}, which are usually optimized by genetic programming \citep{koza1992genetic}. 

Further, much effort in current studies is to research the benefits of model ensembles, which combine several distinct models \citep{goel2007ensemble,muller2014influence,Frie16b}. 
The goal is to create a sophisticated predictor that surpasses the performance of a single model. 
A well-known example is random forest regression\citep{freund1997decision}, which uses \emph{bagging} to fit a large number of decision trees \citep{breiman2001random}. 
We regard ensemble modeling as the state of the art of current research, as they can combine the advantages of different models to generate outstanding results in both classification and regression. 
The drawback of these ensemble methodologies is that they are computationally expensive and pose a severe problem concerning efficient model selection, evaluation, and combination.

\subsubsection{Surrogate-based Algorithms}

Surrogate-based algorithms explicitly utilize a global approximation surrogate in their optimization cycle by following the concept of \emph{efficient global optimization} (EGO) \citep{jones1998efficient} and Bayesian Optimization (BO) \citep{mockus1974bayesian,mockus1994application,mockus2012bayesian}. 
They are either fixed algorithms designed around a specific model, such as \emph{Kriging}, or algorithmic frameworks with a choice of possible surrogates and optimization methods \emph{sequential parameter optimization} \citep{bartz2005sequential,bartz2010spot}.
The basis for our descriptions of surrogate-based algorithms is mainly EGO, and it is to note that the terminology of BO differs partly from our utilized terminology.

A general surrogate-based algorithm can be described as follows: 
\begin{enumerate}
\item The initialization is done by sampling the objective function at $k$ positions with $\vec{y}_{i}=f_1(\vec{x}_i)$ or $\vec{y}_{i}=f_2(\vec{x}_i),1 \le i \le k$ to generate a set of observations 
${\cal{D}}_{t}=\{(\vec{x}_i,\vec{y}_i), 1 \le i \le k \}$. The sampling design plan is commonly selected according to the surrogate.
\item Selecting a suitable surrogate. The selection of the correct surrogate type can be a computational demanding step in the optimization process, as often no prior information indicating the best type is available. 
\item Constructing the surrogate $s(\vec{x})$ using the observations. 
\item Utilizing the surrogate $s(\vec{x})$ to predict $n$ new promising candidates $\{\vec{x}^*_{1:n}\}$, e.g., by optimization of the \emph{infill} function with a suitable algorithm. 
For example, it is reasonable to use algorithms that require a large number of evaluations as the surrogate itself is (comparatively) very cheap to evaluate. 
\item Evaluating the new candidates with the objective function $y^*_i=f(\vec{x}^*_{i}), 1 \le i \le n$.
\item If the stopping criterium is not met: \\
Updating the surrogate with the new observations ${\cal{D}}_{t+1} ={\cal{D}}_{t} \cup \{(\vec{x}^*_i, y^*_i),1 \le i \le n\}$, and repeating the optimization cycle (4.-6.) 
\end{enumerate}

For the initialization, the model building requires a suitable sampling of the search space.
The initial sampling has a significant impact on the performance and should be carefully selected. 
Thus, the initialization commonly uses candidates following different information criteria and suitable experimental designs.
For example,  it is common to built linear regression models with factorial designs and preferably couple Gaussian process models with space-filling designs, such as \emph{Latin hypercube sampling} \citep{montgomery1984design,sacks1989design}. 

The generation has two aspects: the first is the choice of surrogate itself, as it is used to find a new candidate. 
The accuracy of a surrogate strongly relies on the selection of the correct model type to approximate the objective function. 
By selecting a particular surrogate, the user makes certain assumptions regarding the characteristics of the objective function, i.e., modality, continuity, and smoothness \citep{forrester2009recent}. 
Most surrogates are selected to provide continuous, low-modal, and smooth landscapes, which renders the optimization process computational inexpensive and straightforward. 
The second aspect is the optimizer which variates the candidates for the search on the surrogate and the approximated fitness landscape.
As the surrogates are often fast to evaluate, exhaustive \emph{exact} search strategies, such as \emph{branch and bound} in EGO \citet{jones1998efficient} or multi-start hill-climbers, are often utilized, but it is also common to use sophisticated population-based algorithms.

The surrogate prediction for the expected best solution is the basis of the selection of the next candidate solution.
Instead of a simple mean fitness prediction, it is common to define an infill criterion or acquisition function.
Typical choices include the probability of improvement~\citep{kushner1964new}, expected improvement~\citep{jones1998efficient} and confidence bounds~\citep{cox1997sdo}. 
Expected improvement is a common infill criterion because it is a balance of exploration and exploitation by utilizing both the predicted best mean value of the model, as well as the model uncertainty.
The optimization of this infill criterion then selects the candidate.
Typically, in each iteration for evaluation and the model update, the algorithm selects only a single candidate.
Multi-infill selection strategies are also possible.

Surrogate-based algorithms include a large number of control elements, starting with necessary components of such an algorithm,
including the initialization strategy, the choice of surrogate and optimizer. 
Particular, the infill criteria, as part of the selection strategy, has an enormous impact on the performance.
Even for a fixed algorithm, the amount of (required) control is extensive. 
The most important are the model parameters of the surrogate. 

\subsubsection{Surrogate-assisted Algorithms}
\emph{Surrogate-assisted} algorithms utilize s search strategy similar to the population class, but employ a surrogate particular in the \emph{selection} step to preselect candidate solutions based on their approximated fitness and \emph{assist} the evolutionary search strategy~\citep{ong2005surrogate,jin2005comprehensive,emmerich2006single,lim2010generalizing,loshchilov2012self}. 
Commonly, only parts of the new candidates are preselected utilizing the surrogate, while another part follows a direct selection and evaluation process.
The generation and selection of a new candidate are thus not based on an optimization of the surrogate landscape, which is the main difference to the surrogate-based algorithms. 
The surrogate can be built on the archive of all solutions, or locally on the current solution candidates.
An overview of surrogate-assisted optimization is given by \citet{jin2011surrogate}, including several examples for real-world applications, or by \citet{haftka2016parallel} and \citet{rehbach2018comparison}, with focus on parallelization. 

\subsection{Hybrid Class: "The Chimera"} 
\label{sec:hybrid}
\begin{analogy}[The Chimera] \ \\
A chimera is an individual, which is a mixture, composition, or crossover of other individuals. 
It is an entirely new being formed out of original parts from existing species and utilizes their strengths to be versatile. 
\end{analogy}
We describe the combinations of algorithms or their components as the hybrid class. 
Hyperheuristic algorithms also belong to this class.
Overviews of hybrid algorithms were presented by \citet{talbi2002taxonomy,blum2011hybrid,burke2013hyper}
There are two kinds of hybrid algorithms:
\begin{enumerate}
\item \emph{Predetermined Hybrids} (Section \ref{sec:prehybrid}) have a fixed algorithm design, which is composed of certain algorithms or their components.
\item \emph{Automated Hybrids} (Section \ref{sec:authybrid}) use optimization or machine learning to search for an optimal algorithm design or composition. 
\end{enumerate}

\subsubsection{Predetermined Hybrid Algorithms}
\label{sec:prehybrid}
The search strategies of this class improve or tackle algorithm weaknesses or amplify their strengths.  
The algorithms are often given distinctive roles of exploration and exploitation, as they are combinations of an explorative global search method paired with a local search algorithm. 
For example, population-based algorithms with remarkable exploration abilities pair with local algorithms with fast convergence. 
This approach gives some benefits, as the combined algorithms can be adapted or tuned to fulfill their distinct tasks. 
Also well known are \emph{Memetic algorithms}, as defined by \citet{moscato1989evolution}, which are a class of search methods that combine population-based algorithms with local hill-climbers. 
An extensive overview of memetic algorithms is given by \citet{molina2010memetic}. 
They describe how different hybrid algorithms can be constructed by looking at suitable local search algorithms with particular regard to their convergence abilities. 

\subsubsection{Automated Hybrid Algorithms}
\label{sec:authybrid}
Automated hybrids are a special kind of algorithms, which do not use predetermined search strategies, but a pool of different algorithms or algorithm components, 
where the optimal strategy can be composed of \citep{lindauer2015autofolio,bezerra2014automatic}.
Hyperheuristics belong to this class, particularly those that generate new heuristics \citep{burke2010classification}.
Automated algorithm selection tries to find the most suitable algorithm for a specific problem based on machine learning and problem information, such as explorative landscape analysis \citep{kerschke2017automated}.
Instead of selecting individual algorithms, it is tried to select different types of operators for, e.g., generation or selection, to automatically compose a new, best-performing algorithm for a particular problem.
Similar to our defined elements, search operators or components of algorithms are identified, extracted, and then again combined to a new search strategy. 

Generation, variation, and selection focus in this class on algorithm components, instead of candidate solutions. 
The search strategies on this upper level are similar to the presented algorithms; hence, for example, evolutionary or Bayesian techniques are common \citep{guo2003bayesian,vanrijn2017configuration}.

\subsection{Taxonomy: Summary and Examples}

\begin{figure}[ht]
\centering
\includegraphics[width= 1.0\columnwidth] {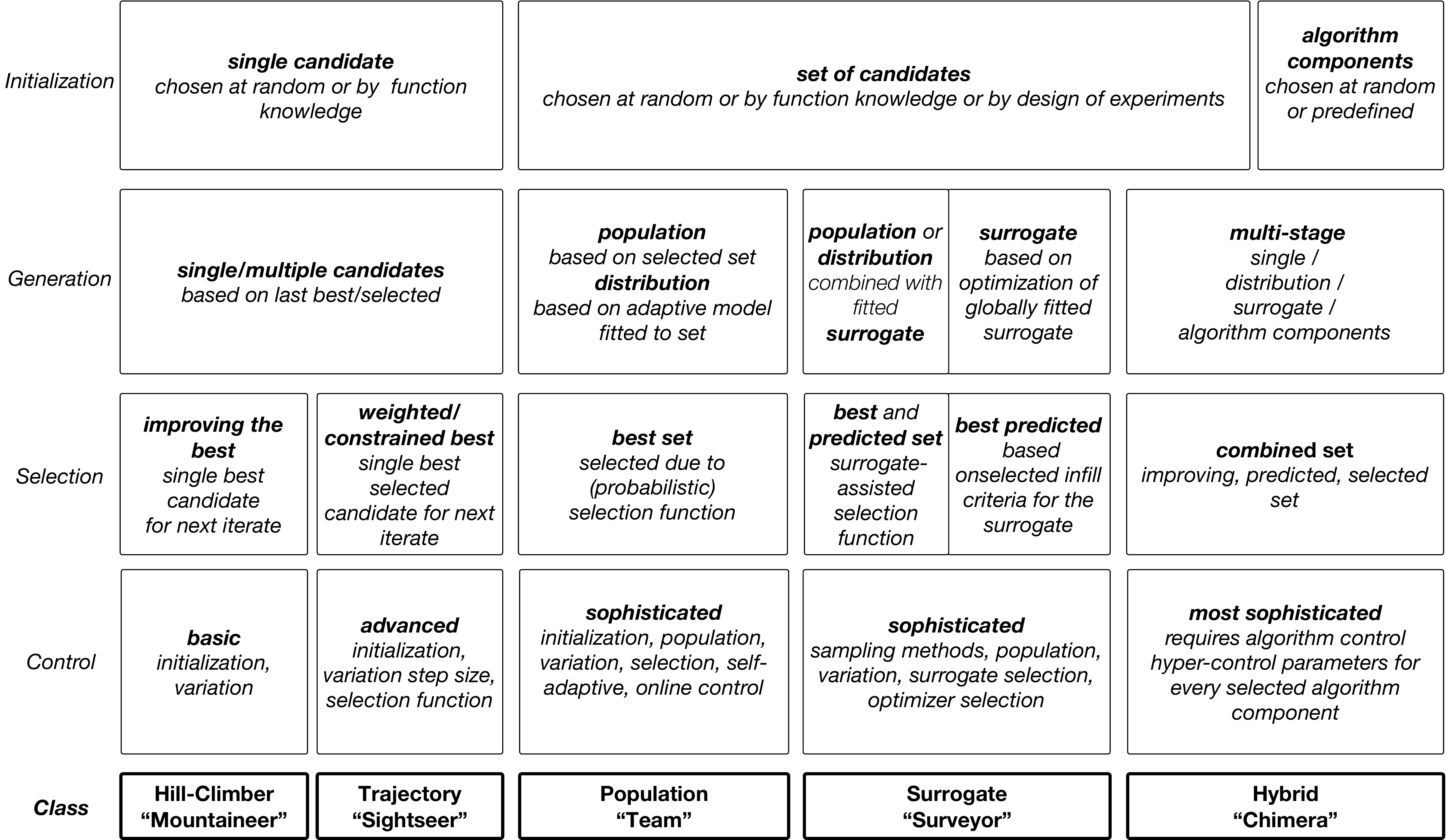}
\caption{Overview of defining algorithm features per search element and class. Overlapping features indicate the close connection between the individual classes.}
\label{fig:classFeatures}
\end{figure} 

Figure \ref{fig:classFeatures} illustrates an overview of all classes and connected features.
It outlines initialization, generation, selection, and control of the individual components for each class.
The algorithm features are partly distinct and define their class, while others are shared.
The figure shows the strong relationship between the algorithm classes; for example, the hill-climbing and trajectory class share similar characteristics.
The algorithms of these classes are similar in their search strategy and built upon each other. 

Table \ref{tab:classExamples} describes examples for each of the defined algorithm classes and outlines their specific features. 
The table is not intended to present a complete overview, instead, for each class and subclass, at least one example is given to 
present the working scheme.
Other algorithms can be easily classified utilizing the scheme presented in Figure \ref{fig:classFeatures}.

\begin{table}[hb]
\caption{Summary of example algorithms for each class of the presented taxonomy}
\label{tab:classExamples}
\begin{tabular}{p{4,5cm}|p{2cm}|p{4cm}}
name                				
& class & specifics                                      \\ \hline
\textbf{1$+$1-Evolution Strategy} \newline \citep{rechenberg1973evolutionsstrategie,schwefel1977numerische}     	
& hill-climber 						& probabilistic, self-adaptive mutation rates   \\  \hline
\textbf{L-BFGS} \citep{liu1989limited}, \newline \textbf{CG} \citep{fletcher1976conjugate} 						
& hill-climber 							& approximating gradient  \\  \hline          
\textbf{Variable Neighborhood Search} \newline \citep{hansen2003variable} 	
& trajectory \newline  (systematic)  			& separation of search space in individually searched sub-spaces   \\ \hline
\textbf{Tabu Search} \citep{glover1989tabu,siarry1997fitting}				
& trajectory \newline  (systematic) 		 		& Tabu-list of search restricted solutions/areas   \\ \hline
\textbf{Simulated Annealing} \newline \citep{kirkpatrick1983optimization}		
& trajectory \newline  (exploring) 				&  control variable: temperature to define exploration strength   \\ \hline
\textbf{Evolutionary Algorithms} \citep{back1996evolutionary,eiben2003introduction,eiben2015introduction}		
& population \newline   (classic)  			& different variation and selection strategies, general framework  \\ \hline
\textbf{Particle Swarm Optimization} \newline \citep{kennedy1995particle}	
& population \newline   (classic)  			& exploration and exploitation strategy based on behavior in swarms \\ \hline
\textbf{Covariance Matrix Adaption - ES} \newline \citep{hansen2003reducing}					
& population \newline   (model)     & self-adaptive, pre-defined parameters, strong performance  \\ \hline
\textbf{Ant Colony Optimization} \newline \citep{dorigo2006ant} 	
& population \newline   (model)  & uses Gaussian process models, based on the behavior of ants \\ \hline
\textbf{Efficient Global Optimization} \newline \citep{jones1998efficient}
& surrogate \newline  (based) & kriging models, expected improvement, high evaluation efficiency   \\ \hline
\textbf{Bayesian Optimization} \newline \citep{mockus1974bayesian,mockus1994application,mockus2012bayesian}
& surrogate \newline (based)  & general framework, high evaluation efficiency  \\ \hline
\textbf{Surrogate-Assisted EAs} \newline \citep{ong2005surrogate,lim2010generalizing}
& surrogate \newline (assisted)  & general framework, assisting evolutionary algorithms with surrogates  \\ \hline
\textbf{Memetic Algorithms} \newline \citep{moscato1989evolution}   		
& hybrid \newline (predetermined) 					 	& combining EAs with hill-climber class unimodal search algorithms \\ \hline 
\textbf{Hyperheuristics} \newline  \citep{burke2003hyper}				
& hybrid \newline (automated)	& automatic selection of entire heuristics or individual search operators\\
\end{tabular}
\end{table}
\section{Algorithm Selection Guideline}
\label{sec:algselect}

\begin{figure}[hb]
\centering
\includegraphics[width= 1.0\columnwidth] {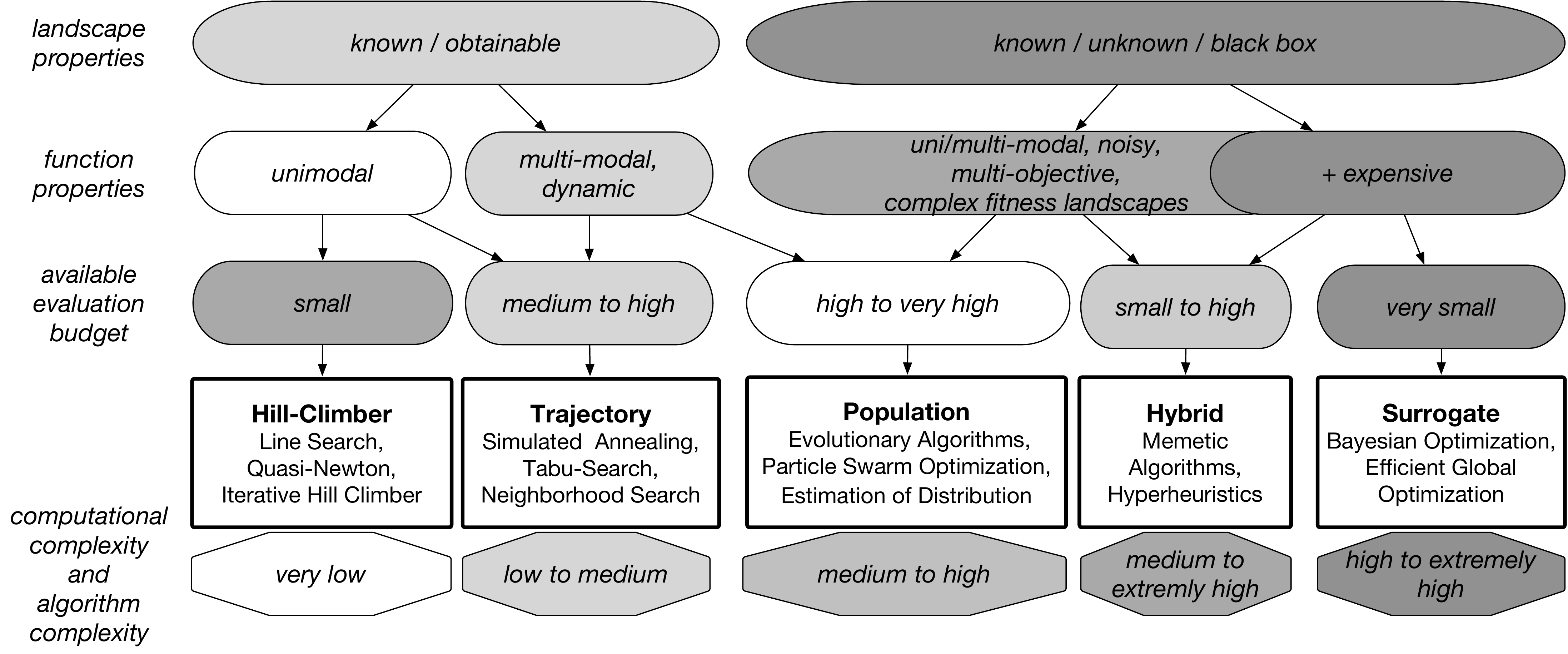}
\caption{Algorithm selection guideline. The figure connects landscape and function properties, as well as the available budget to a suitable algorithm class and outlines their computational complexity.}
\label{fig:classSelection}
\end{figure}

The selection of the best-suited algorithm poses a common problem for practitioners, even if experienced.
In general, it is nearly impossible to predict the performance of any algorithm for a new, unknown problem.
We thus recommend first to gather all available information about the problem if confronted with a new optimization problem.
The features of a problem can be an excellent guideline to select at least an adequate algorithm class, where the users' choice and experience can select a concrete implementation.
Essential features include available information about the fitness landscape and general information about the available resources, both in terms of available evaluations and computational resources.
To help with the selection, we developed a small decision graph which builds upon these significant features. 
The graph is outlined in Figure \ref{fig:classSelection}.

A \emph{hill-climbing} algorithm is in the first place suitable for unimodal functions or to exploit local optima. 
It can be applied for global optimization to multimodal landscapes if an adequate multi-start strategy is employed. 
These multi-start strategies typically demand a high number of function evaluations and are only reasonable to be used for problems with relatively cheap objective functions.

\emph{Exploring trajectory} algorithms are suitable for searches in unimodal and low multimodal problems. 
As they do not rely on stored information of former iterations during their search, they are also an excellent choice to handle dynamic objective functions\citep{carson1997simulation,corana1987minimizing,faber2005dynamic}.
However, the rather simplistic utilization of exploited global information renders them not efficient for challenging and expensive optimization problems. 
Moreover, the control parameters have a significant effect on the performance of these algorithms. We thus advise to tune them in an offline or online fashion.

The central concept of \emph{systematic trajectory} algorithms is to use the information of evaluated solutions and to direct the search to former unknown regions to avoid early convergence to a non-global optimum. 
The strategic use of sub-spaces allows precise control of exploration and exploitation and mainly ensures a high level of exploration. 
They include a large number of parameters, such as the number or size of sub-spaces, which makes them very vulnerable to false setups and less good universal solvers. 
If correctly tuned, algorithms from this class are suitable and efficient for multimodal problems. 
Algorithms using a pre-defined separation of the search space, such as VNS, can utilize domain knowledge for the initial definition of the sub-spaces. 
This pre-defined separation renders them useful for problems where the region of the global optimum is roughly known, but not particularly suitable for black-box problems.

\emph{Population-based} algorithms are very flexible in their implementation and adaptable by tuning. 
They are robust and suitable to solve a large class of problems, including multimodal, multi-objective, dynamic and black-box problems, even with noise or discontinuities in the fitness landscape \citep{jin2005evolutionary,marler2004survey}. 
Further, they have successfully been applied to a large number of different industrial problems \citep{fleming2002evolutionary} but typically require a relatively large number of function evaluations to converge. 
This inefficiency makes them not the first choice for expensive problems, where the number of evaluations is sharply limited. 
Different mechanisms and strategies for controlling the balance between exploration and exploitation cause flexibility and robustness. 
A good overview is presented in the survey by \citet{vcrepinvsek2013exploration}. 
The detailed survey classifies the different available evolutionary approaches and presents an intensive discussion on which mechanisms influence exploration and exploitation. 
Theoretical aspects of evolutionary operators are discussed by \citet{beyer2013theory}. 
\emph{Parameter tuning and control} influence the performance of EAs, e.g., by the setting of population size, mutation strength, and selection probability.  
An extensive overview of the different on- and offline tuning approaches for parameter control in EAs was published by \citet{eiben1999parameter}.  
Further common strategies of controlling exploration and exploitation and multimodal optimization are so-called niching strategies, which utilize sub-populations to maintain the diversity of the population, investigate several regions of the search space in parallel or conduct defined tasks of exploring and exploiting \citep{shir2005niching,filipiak2014infeasibility}.

One step further, \emph{model-based} algorithms try to combine the benefits of statistical models and their capability of storing and processing information with population-based search operators.
They are high-level metaheuristics and advanced EAs which intended to be flexible, robust, and applicable to a large class of problems, particularly those with unknown function properties. 
This generalizability makes them very successful in popular black-box benchmarks \citep{hansen2010comparing}.
For example, the design of the CMA-ES seeks to make the algorithm performance robust and not dependent on the objective function or tuning. 
The various \emph{control parameters} of the algorithm were pre-defined based on theoretical aspects and practical benchmarks.

Surrogate-based algorithms were created to solve expensive problems with the help of an expensive computational surrogate.
Their focus on high evaluation efficiency renders them particularly suitable for problems were only a small number of function evaluations are possible,
such as real-world optimization or physical simulations. 
The high computational complexity of the surrogate fitting and prediction process renders them not applicable to problems that require fast results or a massive amount of function evaluations.
Moreover, solving high-dimensional problems requires unique surrogate models.
In general, the selection of an adequate model, experimental design, and optimizer requires both domain knowledge and expertise. 
\citet{forrester2009recent} and \citet{bartz2017model} give overviews of surrogate-based optimization, different surrogate models and infill criteria and match surrogates to problem classes and give hints about their applicability. 
Surrogate-based optimization was successfully applied to different applications, including expensive optimization problems \citep{lizotte2008practical,khan2002multi} and machine learning \citep{snoek2012practical,swersky2013multi,Stor18c}
Surrogate-assisted optimization is more flexible, as it combines the strength of population-based algorithms with the evaluation efficiency of the surrogate.

Hybrid algorithms apply to a large class of problems, dependent on the origin of their algorithms or components. 
The most recent algorithms from this hybrid class search for the best algorithm design and composition automatically, even at different problem stages or search stages.
The method of automatic algorithm selection has shown to be able to outperform a single algorithm on a set of benchmark functions.
However, this procedure also requires either a large amount of function knowledge or a large evaluation budget and computation time.
Their immense complexity includes the risk that the automatic composition does not lead to improved performance, due to the problematic balancing and required tuning of the distinct algorithms.
Further, their sophisticated search strategies with a large number of control parameters can make them difficult to tune.
For the automatic algorithm selection, numerous operators influence the convergence behavior, and the search strategy itself becomes a black-box that is challenging to comprehend.

In summary, hill-climbing algorithms are only suitable for unimodal problems (or need to utilize defined multi-start strategies) but are very fast and general simple to implement and apply. 
Trajectory class algorithms can tackle multimodal problems, but often require manual setup or meta-optimization of their control parameters to work well, which again requires either function knowledge or additional resources.
If the problem does not provide much information about the underlying function and thus the number of optima, we recommend utilizing algorithms from the population, surrogate or hybrid class, as they apply to several function characteristics.
The population class provides algorithms that have shown to deliver robust performance over a wide range of problems.
Furthermore, we recommend using surrogates if the real objective function is very costly, and only small numbers of evaluations are possible. 
If they are applicable, they can significantly improve the evaluation efficiency and also provide further problem knowledge by accessing the fitted surrogate model statistics.
If optimization speed and computational complexity of the algorithm itself have a priority, we again recommend population-based algorithms, particular modern evolutionary algorithms, and model-based algorithms.
The hybrid class is unique, as it is presumably the most versatile, but also the least specific, as it includes all kinds of algorithm combinations. 
Also, automatic hybrids may require a large number of function evaluations to find a suitable algorithm design.

\section{Concluding Remarks}
In this work, we presented a new comprehensive overview of global optimization algorithms by creating a new taxonomy.
We set a particular focus on covering a broad range of optimization algorithms, including surrogates, metaheuristics, and algorithm combinations, because existing taxonomies do not well cover these. 

Based on a generalized algorithm scheme we defined four characteristic elements of optimization algorithms, 
i.e., how they initialize, generate and select solutions, how these solutions are evaluated and finally, how these algorithms can be parametrized and controlled. 
With these elements, we were able to create a generalized view on optimization algorithms by identifying their specific components. 
These components were then used to divide algorithms in the \emph{hill-climber}, \emph{trajectory}, \emph{population}, \emph{surrogate} and \emph{hybrid} class and to identify similarities and differences
in their search strategies. 

We can conclude that most algorithms and algorithm classes have a close connection and share similar components, operators, and a large part of their search strategies. 
Current research for the automated design of algorithms builds upon this fact. 
It generalizes algorithms by breaking them down to their components and again combining these components to algorithms.
This design process can be fully automatic, selecting components based on known features of the problem.
Recent research aims in this direction \citep{lindauer2015autofolio,kerschke2017automated,bezerra2014automatic,vanrijn2017configuration}. 
These automatic methods explicitly need classifications of algorithms to build their design spaces of suitable algorithms classes.
Our classes and identified elements can be utilized to create a design scheme for each class. 
For example, employ an algorithm generator for hill-climbers or population-based algorithms.
Our taxonomy is thus particularly useful for these automatic methods, as it includes a broad set of current developments in the presented classification scheme. 

Our set of accompanying analogies, based on the human behavior in pathfinding, mainly helps novices to comprehend the fundamentals of the algorithm search strategies. 
For more advanced practitioners, we further outlined an algorithm selection guideline with best practices, which can support them if they face a new problem and have to choose a suitable class of optimization algorithms.

Exciting challenges for future algorithm design arise from problems in two categories. The first is \emph{industry 4.0}. 
Digitalization in manufacturing and engineering, particularly the rapid development of communicating sensors and machines in the field of engineering, requires new designs. 
Suitable optimization algorithms need to be directly included in the production cycle, adapting to generate robust solutions in challenging dynamic environments in an online manner, robustly improving themselves over a long-time period in the field. 
The second category is machine learning techniques, particularly the training of models with a massive amount of parameters, such as deep learning networks. 
They require algorithm designs that combine sampling and computation time efficiency, which is a challenging optimization problem on its own.
Regarding this problem, we further identify a lack in the field of realistic benchmarks, which are related or based on real-world applications and would allow a realistic comparison of different approaches to work in practice.

\section*{Declarations of Interest}
The authors declare that they have no conflict of interest.

\bibliography{Stor18b}

\begin{thebibliography}{134}
\providecommand{\natexlab}[1]{#1}
\providecommand{\url}[1]{{#1}}
\providecommand{\urlprefix}{URL }
\expandafter\ifx\csname urlstyle\endcsname\relax
  \providecommand{\doi}[1]{DOI~\discretionary{}{}{}#1}\else
  \providecommand{\doi}{DOI~\discretionary{}{}{}\begingroup
  \urlstyle{rm}\Url}\fi
\providecommand{\eprint}[2][]{\url{#2}}

\bibitem[{Angeline(1995)}]{angeline1995adaptive}
Angeline PJ (1995) Adaptive and self-adaptive evolutionary computations. In:
  Computational intelligence: a dynamic systems perspective, Citeseer

\bibitem[{Archetti and Schoen(1984)}]{Arch84a}
Archetti F, Schoen F (1984) {A survey on the global optimization problem:
  general theory and computational approaches}. Annals of Operations Research
  1(2):87--110

\bibitem[{Arnold and Beyer(2003)}]{arnold2003comparison}
Arnold DV, Beyer HG (2003) A comparison of evolution strategies with other
  direct search methods in the presence of noise. Computational Optimization
  and Applications 24(1):135--159

\bibitem[{Arnold and Hansen(2012)}]{arnold20121}
Arnold DV, Hansen N (2012) A (1+ 1)-cma-es for constrained optimisation. In:
  Proceedings of the 14th annual conference on Genetic and evolutionary
  computation, ACM, pp 297--304

\bibitem[{Arora et~al.(1995)Arora, Elwakeil, Chahande, and
  Hsieh}]{arora1995global}
Arora J, Elwakeil O, Chahande A, Hsieh C (1995) Global optimization methods for
  engineering applications: a review. Structural optimization 9(3-4):137--159

\bibitem[{Audet(2014)}]{audet2014survey}
Audet C (2014) A survey on direct search methods for blackbox optimization and
  their applications. In: Mathematics Without Boundaries, Springer, pp 31--56

\bibitem[{Augusto and Barbosa(2000)}]{augusto2000symbolic}
Augusto DA, Barbosa HJ (2000) Symbolic regression via genetic programming. In:
  Neural Networks, 2000. Proceedings. Sixth Brazilian Symposium on, IEEE, pp
  173--178

\bibitem[{Back(1996)}]{back1996evolutionary}
Back T (1996) Evolutionary algorithms in theory and practice: evolution
  strategies, evolutionary programming, genetic algorithms. Oxford university
  press

\bibitem[{Back et~al.(1997)Back, Fogel, and Michalewicz}]{back1997handbook}
Back T, Fogel DB, Michalewicz Z (1997) Handbook of evolutionary computation.
  IOP Publishing Ltd.

\bibitem[{Bartz-Beielstein(2010)}]{bartz2010spot}
Bartz-Beielstein T (2010) Spot: An r package for automatic and interactive
  tuning of optimization algorithms by sequential parameter optimization. arXiv
  preprint arXiv:10064645

\bibitem[{Bartz-Beielstein and Zaefferer(2017)}]{bartz2017model}
Bartz-Beielstein T, Zaefferer M (2017) Model-based methods for continuous and
  discrete global optimization. Applied Soft Computing 55:154--167

\bibitem[{Bartz-Beielstein et~al.(2005)Bartz-Beielstein, Lasarczyk, and
  Preu{\ss}}]{bartz2005sequential}
Bartz-Beielstein T, Lasarczyk CW, Preu{\ss} M (2005) Sequential parameter
  optimization. In: Evolutionary Computation, 2005. The 2005 IEEE Congress on,
  IEEE, vol~1, pp 773--780

\bibitem[{Beyer(2013)}]{beyer2013theory}
Beyer HG (2013) The theory of evolution strategies. Springer Science \&
  Business Media

\bibitem[{Bezerra et~al.(2014)Bezerra, L{\'o}pez-Ib{\'a}nez, and
  St{\"u}tzle}]{bezerra2014automatic}
Bezerra LC, L{\'o}pez-Ib{\'a}nez M, St{\"u}tzle T (2014) Automatic design of
  evolutionary algorithms for multi-objective combinatorial optimization. In:
  International Conference on Parallel Problem Solving from Nature, Springer,
  pp 508--517

\bibitem[{Blum et~al.(2011)Blum, Puchinger, Raidl, and Roli}]{blum2011hybrid}
Blum C, Puchinger J, Raidl GR, Roli A (2011) Hybrid metaheuristics in
  combinatorial optimization: A survey. Applied Soft Computing 11(6):4135--4151

\bibitem[{Boussa{\"\i}d et~al.(2013)Boussa{\"\i}d, Lepagnot, and
  Siarry}]{boussaid2013survey}
Boussa{\"\i}d I, Lepagnot J, Siarry P (2013) A survey on optimization
  metaheuristics. Information Sciences 237:82--117

\bibitem[{Breiman(2001)}]{breiman2001random}
Breiman L (2001) Random forests. Machine Learning 45(1):5--32

\bibitem[{Breiman et~al.(1984)Breiman, Friedman, Stone, and
  Olshen}]{breiman1984classification}
Breiman L, Friedman J, Stone CJ, Olshen RA (1984) Classification and regression
  trees. CRC press

\bibitem[{Burke et~al.(2003)Burke, Kendall, Newall, Hart, Ross, and
  Schulenburg}]{burke2003hyper}
Burke E, Kendall G, Newall J, Hart E, Ross P, Schulenburg S (2003)
  Hyper-heuristics: An emerging direction in modern search technology. In:
  Handbook of metaheuristics, Springer, pp 457--474

\bibitem[{Burke et~al.(2010)Burke, Hyde, Kendall, Ochoa, {\"O}zcan, and
  Woodward}]{burke2010classification}
Burke EK, Hyde M, Kendall G, Ochoa G, {\"O}zcan E, Woodward JR (2010) A
  classification of hyper-heuristic approaches. In: Handbook of metaheuristics,
  Springer, pp 449--468

\bibitem[{Burke et~al.(2013)Burke, Gendreau, Hyde, Kendall, Ochoa, {\"O}zcan,
  and Qu}]{burke2013hyper}
Burke EK, Gendreau M, Hyde M, Kendall G, Ochoa G, {\"O}zcan E, Qu R (2013)
  Hyper-heuristics: A survey of the state of the art. Journal of the
  Operational Research Society 64(12):1695--1724

\bibitem[{Carson and Maria(1997)}]{carson1997simulation}
Carson Y, Maria A (1997) Simulation optimization: methods and applications. In:
  Proceedings of the 29th conference on Winter simulation, IEEE Computer
  Society, pp 118--126

\bibitem[{Chelouah and Siarry(2000)}]{chelouah2000tabu}
Chelouah R, Siarry P (2000) Tabu search applied to global optimization.
  European journal of operational research 123(2):256--270

\bibitem[{Coello(2002)}]{coello2002theoretical}
Coello CAC (2002) Theoretical and numerical constraint-handling techniques used
  with evolutionary algorithms: a survey of the state of the art. Computer
  methods in applied mechanics and engineering 191(11):1245--1287

\bibitem[{Collobert and Weston(2008)}]{collobert2008unified}
Collobert R, Weston J (2008) A unified architecture for natural language
  processing: Deep neural networks with multitask learning. In: Proceedings of
  the 25th international conference on Machine learning, ACM, pp 160--167

\bibitem[{Corana et~al.(1987)Corana, Marchesi, Martini, and
  Ridella}]{corana1987minimizing}
Corana A, Marchesi M, Martini C, Ridella S (1987) Minimizing multimodal
  functions of continuous variables with the ``simulated annealing'' algorithm.
  ACM Transactions on Mathematical Software (TOMS) 13(3):262--280

\bibitem[{Cowling et~al.(2000)Cowling, Kendall, and
  Soubeiga}]{cowling2000hyperheuristic}
Cowling P, Kendall G, Soubeiga E (2000) A hyperheuristic approach to scheduling
  a sales summit. In: International Conference on the Practice and Theory of
  Automated Timetabling, Springer, pp 176--190

\bibitem[{Cowling et~al.(2002)Cowling, Kendall, and
  Soubeiga}]{cowling2002hyperheuristics}
Cowling P, Kendall G, Soubeiga E (2002) Hyperheuristics: A tool for rapid
  prototyping in scheduling and optimisation. In: Workshops on Applications of
  Evolutionary Computation, Springer, pp 1--10

\bibitem[{Cox and John(1997)}]{cox1997sdo}
Cox DD, John S (1997) Sdo: A statistical method for global optimization.
  Multidisciplinary design optimization: state of the art pp 315--329

\bibitem[{{\v{C}}repin{\v{s}}ek et~al.(2013){\v{C}}repin{\v{s}}ek, Liu, and
  Mernik}]{vcrepinvsek2013exploration}
{\v{C}}repin{\v{s}}ek M, Liu SH, Mernik M (2013) Exploration and exploitation
  in evolutionary algorithms: a survey. ACM Computing Surveys (CSUR) 45(3):35

\bibitem[{Dorigo et~al.(2006)Dorigo, Birattari, and Stutzle}]{dorigo2006ant}
Dorigo M, Birattari M, Stutzle T (2006) Ant colony optimization. IEEE
  computational intelligence magazine 1(4):28--39

\bibitem[{Eiben and Smith(2003)}]{eiben2003introduction}
Eiben AE, Smith JE (2003) Introduction to evolutionary computing, vol~53.
  Springer

\bibitem[{Eiben and Smith(2015)}]{eiben2015introduction}
Eiben AE, Smith JE (2015) Introduction to Evolutionary Computing, 2nd edn.
  Springer

\bibitem[{Eiben et~al.(1999)Eiben, Hinterding, and
  Michalewicz}]{eiben1999parameter}
Eiben AE, Hinterding R, Michalewicz Z (1999) Parameter control in evolutionary
  algorithms. Evolutionary Computation, IEEE Transactions on 3(2):124--141

\bibitem[{Emmerich et~al.(2006)Emmerich, Giannakoglou, and
  Naujoks}]{emmerich2006single}
Emmerich MT, Giannakoglou KC, Naujoks B (2006) Single-and multiobjective
  evolutionary optimization assisted by gaussian random field metamodels. IEEE
  Transactions on Evolutionary Computation 10(4):421--439

\bibitem[{Faber et~al.(2005)Faber, Jockenh{\"o}vel, and
  Tsatsaronis}]{faber2005dynamic}
Faber R, Jockenh{\"o}vel T, Tsatsaronis G (2005) Dynamic optimization with
  simulated annealing. Computers \& chemical engineering 29(2):273--290

\bibitem[{Filipiak and Lipinski(2014)}]{filipiak2014infeasibility}
Filipiak P, Lipinski P (2014) Infeasibility driven evolutionary algorithm with
  feed-forward prediction strategy for dynamic constrained optimization
  problems. In: Applications of Evolutionary Computation, Springer, pp 817--828

\bibitem[{Flasch et~al.(2010)Flasch, Mersmann, and
  Bartz-Beielstein}]{2010flasch}
Flasch O, Mersmann O, Bartz-Beielstein T (2010) Rgp: An open source genetic
  programming system for the r environment. In: Proceedings of the 12th Annual
  Conference Companion on Genetic and Evolutionary Computation, ACM, New York,
  NY, USA, GECCO '10, pp 2071--2072, \doi{10.1145/1830761.1830867},
  \urlprefix\url{http://doi.acm.org/10.1145/1830761.1830867}

\bibitem[{Fleming and Purshouse(2002)}]{fleming2002evolutionary}
Fleming PJ, Purshouse RC (2002) Evolutionary algorithms in control systems
  engineering: a survey. Control engineering practice 10(11):1223--1241

\bibitem[{Fletcher(1976)}]{fletcher1976conjugate}
Fletcher R (1976) Conjugate gradient methods for indefinite systems. In:
  Numerical analysis, Springer, pp 73--89

\bibitem[{Fogel et~al.(1966)Fogel, Owens, and Walsh}]{fogel1966artificial}
Fogel LJ, Owens AJ, Walsh MJ (1966) Artificial intelligence through simulated
  evolution. John Wiley

\bibitem[{Fomin and Kaski(2013)}]{Fomi13a}
Fomin FV, Kaski P (2013) Exact exponential algorithms. Commun ACM 56(3):80--88,
  \doi{10.1145/2428556.2428575},
  \urlprefix\url{http://doi.acm.org/10.1145/2428556.2428575}

\bibitem[{Fonseca et~al.(1993)Fonseca, Fleming et~al.}]{fonseca1993genetic}
Fonseca CM, Fleming PJ, et~al. (1993) Genetic algorithms for multiobjective
  optimization: Formulationdiscussion and generalization. In: Icga, vol~93, pp
  416--423

\bibitem[{Forrester et~al.(2008)Forrester, Sobester, and
  Keane}]{forrester2008engineering}
Forrester A, Sobester A, Keane A (2008) Engineering design via surrogate
  modelling: a practical guide. John Wiley \& Sons, Inc.

\bibitem[{Forrester and Keane(2009)}]{forrester2009recent}
Forrester AI, Keane AJ (2009) Recent advances in surrogate-based optimization.
  Progress in Aerospace Sciences 45(1):50--79

\bibitem[{Freund and Schapire(1997)}]{freund1997decision}
Freund Y, Schapire RE (1997) A decision-theoretic generalization of on-line
  learning and an application to boosting. Journal of computer and system
  sciences 55(1):119--139

\bibitem[{Friese et~al.(2016)Friese, Bartz-Beielstein, and Emmerich}]{Frie16b}
Friese M, Bartz-Beielstein T, Emmerich MTM (2016) {Building ensembles of
  surrogates by optimal convex combination}. In: Mernik M, Papa G (eds)
  Bioinspired Optimization Methods and their Applications, pp 131--144

\bibitem[{G{\"a}mperle et~al.(2002)G{\"a}mperle, M{\"u}ller, and
  Koumoutsakos}]{gamperle2002parameter}
G{\"a}mperle R, M{\"u}ller SD, Koumoutsakos P (2002) A parameter study for
  differential evolution. Advances in intelligent systems, fuzzy systems,
  evolutionary computation 10:293--298

\bibitem[{Glover(1989)}]{glover1989tabu}
Glover F (1989) Tabu search---part i. ORSA Journal on computing 1(3):190--206

\bibitem[{Goel et~al.(2007)Goel, Haftka, Shyy, and Queipo}]{goel2007ensemble}
Goel T, Haftka RT, Shyy W, Queipo NV (2007) Ensemble of surrogates. Structural
  and Multidisciplinary Optimization 33(3):199--216

\bibitem[{Goffe et~al.(1994)Goffe, Ferrier, and Rogers}]{goffe1994global}
Goffe WL, Ferrier GD, Rogers J (1994) Global optimization of statistical
  functions with simulated annealing. Journal of Econometrics 60(1):65--99

\bibitem[{Guo(2003)}]{guo2003bayesian}
Guo H (2003) A bayesian approach for automatic algorithm selection. In:
  Proceedings of the International Joint Conference on Artificial Intelligence
  (IJCAI03), Workshop on AI and Autonomic Computing, Acapulco, Mexico, pp 1--5

\bibitem[{Haftka et~al.(2016)Haftka, Villanueva, and
  Chaudhuri}]{haftka2016parallel}
Haftka RT, Villanueva D, Chaudhuri A (2016) Parallel surrogate-assisted global
  optimization with expensive functions--a survey. Structural and
  Multidisciplinary Optimization 54(1):3--13

\bibitem[{Hansen et~al.(2003)Hansen, M{\"u}ller, and
  Koumoutsakos}]{hansen2003reducing}
Hansen N, M{\"u}ller SD, Koumoutsakos P (2003) Reducing the time complexity of
  the derandomized evolution strategy with covariance matrix adaptation
  (cma-es). Evolutionary computation 11(1):1--18

\bibitem[{Hansen et~al.(2010{\natexlab{a}})Hansen, Auger, Ros, Finck, and
  Po{\v{s}}{\'\i}k}]{hansen2010comparing}
Hansen N, Auger A, Ros R, Finck S, Po{\v{s}}{\'\i}k P (2010{\natexlab{a}})
  Comparing results of 31 algorithms from the black-box optimization
  benchmarking bbob-2009. In: Proceedings of the 12th annual conference
  companion on Genetic and evolutionary computation, ACM, pp 1689--1696

\bibitem[{Hansen and Mladenovic(2003)}]{hansen2003variable}
Hansen P, Mladenovic N (2003) Variable neighbourhood search. Handbook of
  Metaheuristics, Dordrecht, Kluwer Academic Publishers

\bibitem[{Hansen et~al.(2010{\natexlab{b}})Hansen, Mladenovi{\'c}, and
  P{\'e}rez}]{hansen2010variable}
Hansen P, Mladenovi{\'c} N, P{\'e}rez JAM (2010{\natexlab{b}}) Variable
  neighbourhood search: methods and applications. Annals of Operations Research
  175(1):367--407

\bibitem[{Hauschild and Pelikan(2011)}]{hauschild2011introduction}
Hauschild M, Pelikan M (2011) An introduction and survey of estimation of
  distribution algorithms. Swarm and Evolutionary Computation 1(3):111--128

\bibitem[{Haykin(2004)}]{haykin2004comprehensive}
Haykin S (2004) A comprehensive foundation. Neural Networks 2(2004)

\bibitem[{Henderson et~al.(2003)Henderson, Jacobson, and
  Johnson}]{henderson2003theory}
Henderson D, Jacobson SH, Johnson AW (2003) The theory and practice of
  simulated annealing. In: Handbook of metaheuristics, Springer, pp 287--319

\bibitem[{Hinton et~al.(2012)Hinton, Deng, Yu, Dahl, Mohamed, Jaitly, Senior,
  Vanhoucke, Nguyen, Sainath et~al.}]{hinton2012deep}
Hinton G, Deng L, Yu D, Dahl GE, Mohamed Ar, Jaitly N, Senior A, Vanhoucke V,
  Nguyen P, Sainath TN, et~al. (2012) Deep neural networks for acoustic
  modeling in speech recognition: The shared views of four research groups.
  IEEE Signal Processing Magazine 29(6):82--97

\bibitem[{Hinton et~al.(2006)Hinton, Osindero, and Teh}]{hinton2006fast}
Hinton GE, Osindero S, Teh YW (2006) A fast learning algorithm for deep belief
  nets. Neural computation 18(7):1527--1554

\bibitem[{Holland(1992)}]{holland1992adaptation}
Holland JH (1992) Adaptation in natural and artificial systems: an introductory
  analysis with applications to biology, control, and artificial intelligence

\bibitem[{Hornik et~al.(1989)Hornik, Stinchcombe, and
  White}]{hornik1989multilayer}
Hornik K, Stinchcombe M, White H (1989) Multilayer feedforward networks are
  universal approximators. Neural networks 2(5):359--366

\bibitem[{Hu(1992)}]{hu1992tabu}
Hu N (1992) Tabu search method with random moves for globally optimal design.
  International Journal for Numerical Methods in Engineering 35(5):1055--1070

\bibitem[{Hutter et~al.(2011)Hutter, Hoos, and
  Leyton-Brown}]{hutter2011sequential}
Hutter F, Hoos HH, Leyton-Brown K (2011) Sequential model-based optimization
  for general algorithm configuration. LION 5:507--523

\bibitem[{Jin(2005)}]{jin2005comprehensive}
Jin Y (2005) A comprehensive survey of fitness approximation in evolutionary
  computation. Soft Computing-A Fusion of Foundations, Methodologies and
  Applications 9(1):3--12

\bibitem[{Jin(2011)}]{jin2011surrogate}
Jin Y (2011) Surrogate-assisted evolutionary computation: Recent advances and
  future challenges. Swarm and Evolutionary Computation 1(2):61--70

\bibitem[{Jin and Branke(2005)}]{jin2005evolutionary}
Jin Y, Branke J (2005) Evolutionary optimization in uncertain environments-a
  survey. IEEE Transactions on evolutionary computation 9(3):303--317

\bibitem[{Jones(2001)}]{jones2001taxonomy}
Jones DR (2001) A taxonomy of global optimization methods based on response
  surfaces. Journal of global optimization 21(4):345--383

\bibitem[{Jones et~al.(1998)Jones, Schonlau, and Welch}]{jones1998efficient}
Jones DR, Schonlau M, Welch WJ (1998) Efficient global optimization of
  expensive black-box functions. Journal of Global optimization 13(4):455--492

\bibitem[{Kennedy and Eberhart(1995)}]{kennedy1995particle}
Kennedy J, Eberhart R (1995) Particle swarm optimization. In: Neural Networks,
  1995. Proceedings., IEEE International Conference on, IEEE, vol~4, pp
  1942--1948

\bibitem[{Kerschke and Trautmann(2017)}]{kerschke2017automated}
Kerschke P, Trautmann H (2017) Automated algorithm selection on continuous
  black-box problems by combining exploratory landscape analysis and machine
  learning. arXiv preprint arXiv:171108921

\bibitem[{Khan et~al.(2002)Khan, Goldberg, and Pelikan}]{khan2002multi}
Khan N, Goldberg DE, Pelikan M (2002) Multi-objective bayesian optimization
  algorithm. In: Proceedings of the 4th Annual Conference on Genetic and
  Evolutionary Computation, Morgan Kaufmann Publishers Inc., San Francisco, CA,
  USA, GECCO'02, pp 684--684,
  \urlprefix\url{http://dl.acm.org/citation.cfm?id=2955491.2955599}

\bibitem[{Kirkpatrick et~al.(1983)Kirkpatrick, Gelatt, Vecchi
  et~al.}]{kirkpatrick1983optimization}
Kirkpatrick S, Gelatt CD, Vecchi MP, et~al. (1983) Optimization by simulated
  annealing. science 220(4598):671--680

\bibitem[{Kolda et~al.(2003)Kolda, Lewis, and Torczon}]{kolda2003optimization}
Kolda TG, Lewis RM, Torczon V (2003) Optimization by direct search: New
  perspectives on some classical and modern methods. SIAM review 45(3):385--482

\bibitem[{Koza(1992)}]{koza1992genetic}
Koza JR (1992) Genetic programming: on the programming of computers by means of
  natural selection, vol~1. MIT press

\bibitem[{Kushner(1964)}]{kushner1964new}
Kushner HJ (1964) A new method of locating the maximum point of an arbitrary
  multipeak curve in the presence of noise. Journal of Basic Engineering
  86(1):97--106

\bibitem[{Larra{\~n}aga and Lozano(2001)}]{larranaga2001estimation}
Larra{\~n}aga P, Lozano JA (2001) Estimation of distribution algorithms: A new
  tool for evolutionary computation, vol~2. Springer Science \& Business Media

\bibitem[{Leon(1966)}]{leon1966classified}
Leon A (1966) A classified bibliography on optimization. Recent Advances in
  Optimization Techniques, John Wiley \& Sons, Inc, New York and London pp
  599--649

\bibitem[{Lewis et~al.(2000)Lewis, Torczon, and Trosset}]{lewis2000direct}
Lewis RM, Torczon V, Trosset MW (2000) Direct search methods: then and now.
  Journal of computational and Applied Mathematics 124(1):191--207

\bibitem[{Lim et~al.(2010)Lim, Jin, Ong, and Sendhoff}]{lim2010generalizing}
Lim D, Jin Y, Ong YS, Sendhoff B (2010) Generalizing surrogate-assisted
  evolutionary computation. IEEE Transactions on Evolutionary Computation
  14(3):329

\bibitem[{Lindauer et~al.(2015)Lindauer, Hoos, Hutter, and
  Schaub}]{lindauer2015autofolio}
Lindauer M, Hoos HH, Hutter F, Schaub T (2015) Autofolio: An automatically
  configured algorithm selector. Journal of Artificial Intelligence Research
  53:745--778

\bibitem[{Liu and Nocedal(1989)}]{liu1989limited}
Liu DC, Nocedal J (1989) On the limited memory bfgs method for large scale
  optimization. Mathematical programming 45(1-3):503--528

\bibitem[{Lizotte(2008)}]{lizotte2008practical}
Lizotte DJ (2008) Practical bayesian optimization. University of Alberta

\bibitem[{Lobo et~al.(2007)Lobo, Lima, and Michalewicz}]{lobo2007parameter}
Lobo F, Lima CF, Michalewicz Z (2007) Parameter setting in evolutionary
  algorithms, vol~54. Springer Science \& Business Media

\bibitem[{Locatelli(2002)}]{locatelli2002simulated}
Locatelli M (2002) Simulated annealing algorithms for continuous global
  optimization. In: Handbook of global optimization, Springer, pp 179--229

\bibitem[{L{\'o}pez-Ib{\'a}{\~n}ez et~al.(2016)L{\'o}pez-Ib{\'a}{\~n}ez,
  Dubois-Lacoste, C{\'a}ceres, Birattari, and St{\"u}tzle}]{lopez2016irace}
L{\'o}pez-Ib{\'a}{\~n}ez M, Dubois-Lacoste J, C{\'a}ceres LP, Birattari M,
  St{\"u}tzle T (2016) The irace package: Iterated racing for automatic
  algorithm configuration. Operations Research Perspectives 3:43--58

\bibitem[{Loshchilov et~al.(2012)Loshchilov, Schoenauer, and
  Sebag}]{loshchilov2012self}
Loshchilov I, Schoenauer M, Sebag M (2012) Self-adaptive surrogate-assisted
  covariance matrix adaptation evolution strategy. In: Proceedings of the 14th
  annual conference on Genetic and evolutionary computation, ACM, pp 321--328

\bibitem[{Marler and Arora(2004)}]{marler2004survey}
Marler RT, Arora JS (2004) Survey of multi-objective optimization methods for
  engineering. Structural and multidisciplinary optimization 26(6):369--395

\bibitem[{Marsden et~al.(2004)Marsden, Wang, Dennis~Jr, and
  Moin}]{marsden2004optimal}
Marsden AL, Wang M, Dennis~Jr JE, Moin P (2004) Optimal aeroacoustic shape
  design using the surrogate management framework. Optimization and Engineering
  5(2):235--262

\bibitem[{McKay et~al.(1995)McKay, Willis, and Barton}]{mckay1995using}
McKay B, Willis MJ, Barton GW (1995) Using a tree structured genetic algorithm
  to perform symbolic regression. In: Genetic Algorithms in Engineering
  Systems: Innovations and Applications, 1995. GALESIA. First International
  Conference on (Conf. Publ. No. 414), IET, pp 487--492

\bibitem[{Mercer and Sampson(1978)}]{mercer1978adaptive}
Mercer RE, Sampson J (1978) Adaptive search using a reproductive meta-plan.
  Kybernetes 7(3):215--228

\bibitem[{Mladenovi{\'c} et~al.(2008)Mladenovi{\'c}, Dra{\v{z}}i{\'c},
  Kova{\v{c}}evic-Vuj{\v{c}}i{\'c}, and
  {\v{C}}angalovi{\'c}}]{mladenovic2008general}
Mladenovi{\'c} N, Dra{\v{z}}i{\'c} M, Kova{\v{c}}evic-Vuj{\v{c}}i{\'c} V,
  {\v{C}}angalovi{\'c} M (2008) General variable neighborhood search for the
  continuous optimization. European Journal of Operational Research
  191(3):753--770

\bibitem[{Mockus(1974)}]{mockus1974bayesian}
Mockus J (1974) On bayesian methods for seeking the extremum. In: Proceedings
  of the IFIP Technical Conference, Springer-Verlag, pp 400--404

\bibitem[{Mockus(1994)}]{mockus1994application}
Mockus J (1994) Application of bayesian approach to numerical methods of global
  and stochastic optimization. Journal of Global Optimization 4(4):347--365

\bibitem[{Mockus(2012)}]{mockus2012bayesian}
Mockus J (2012) Bayesian approach to global optimization: theory and
  applications, vol~37. Springer Science \& Business Media

\bibitem[{Molina et~al.(2010)Molina, Lozano, Garc{\'\i}a-Mart{\'\i}nez, and
  Herrera}]{molina2010memetic}
Molina D, Lozano M, Garc{\'\i}a-Mart{\'\i}nez C, Herrera F (2010) Memetic
  algorithms for continuous optimisation based on local search chains.
  Evolutionary Computation 18(1):27--63

\bibitem[{Montgomery et~al.(1984)Montgomery, Montgomery, and
  Montgomery}]{montgomery1984design}
Montgomery DC, Montgomery DC, Montgomery DC (1984) Design and analysis of
  experiments, vol~7. Wiley New York

\bibitem[{Moscato et~al.(1989)}]{moscato1989evolution}
Moscato P, et~al. (1989) On evolution, search, optimization, genetic algorithms
  and martial arts: Towards memetic algorithms. Caltech concurrent computation
  program, C3P Report 826:1989

\bibitem[{M{\"u}ller and Shoemaker(2014)}]{muller2014influence}
M{\"u}ller J, Shoemaker CA (2014) Influence of ensemble surrogate models and
  sampling strategy on the solution quality of algorithms for computationally
  expensive black-box global optimization problems. Journal of Global
  Optimization 60(2):123--144

\bibitem[{Naujoks et~al.(2005)Naujoks, Beume, and Emmerich}]{naujoks2005multi}
Naujoks B, Beume N, Emmerich M (2005) Multi-objective optimisation using
  s-metric selection: Application to three-dimensional solution spaces. In:
  Evolutionary Computation, 2005. The 2005 IEEE Congress on, IEEE, vol~2, pp
  1282--1289

\bibitem[{Nelder and Mead(1965)}]{nelder1965simplex}
Nelder JA, Mead R (1965) A simplex method for function minimization. The
  computer journal 7(4):308--313

\bibitem[{Neumaier(2004)}]{neumaier2004complete}
Neumaier A (2004) Complete search in continuous global optimization and
  constraint satisfaction. Acta numerica 13:271--369

\bibitem[{Ong et~al.(2005)Ong, Nair, Keane, and Wong}]{ong2005surrogate}
Ong YS, Nair P, Keane A, Wong K (2005) Surrogate-assisted evolutionary
  optimization frameworks for high-fidelity engineering design problems. In:
  Knowledge Incorporation in Evolutionary Computation, Springer, pp 307--331

\bibitem[{Pearl(1985)}]{pearl1985heuristics}
Pearl J (1985) Heuristics. intelligent search strategies for computer problem
  solving. The Addison-Wesley Series in Artificial Intelligence, Reading, Mass:
  Addison-Wesley, 1985, Reprinted version

\bibitem[{Queipo et~al.(2005)Queipo, Haftka, Shyy, Goel, Vaidyanathan, and
  Tucker}]{queipo2005surrogate}
Queipo NV, Haftka RT, Shyy W, Goel T, Vaidyanathan R, Tucker PK (2005)
  Surrogate-based analysis and optimization. Progress in aerospace sciences
  41(1):1--28

\bibitem[{Rechenberg(1973)}]{rechenberg1973evolutionsstrategie}
Rechenberg I (1973) Evolutionsstrategie--Optimierung technischer Systeme nach
  Prinzipien der biologischen Evolution. Frommann-Holzboog

\bibitem[{Rechenberg(1994)}]{rechenberg1994}
Rechenberg I (1994) Evolutionsstrategie '94. Frommann-Holzboog

\bibitem[{Rehbach et~al.(2018)Rehbach, Zaefferer, Stork, and
  Bartz-Beielstein}]{rehbach2018comparison}
Rehbach F, Zaefferer M, Stork J, Bartz-Beielstein T (2018) Comparison of
  parallel surrogate-assisted optimization approaches. In: Proceedings of the
  Genetic and Evolutionary Computation Conference, ACM, pp 1348--1355

\bibitem[{van Rijn et~al.(2017)van Rijn, Wang, van Stein, and
  B\"{a}ck}]{vanrijn2017configuration}
van Rijn S, Wang H, van Stein B, B\"{a}ck T (2017) Algorithm configuration data
  mining for cma evolution strategies. In: Proceedings of the Genetic and
  Evolutionary Computation Conference, ACM, New York, NY, USA, GECCO '17, pp
  737--744, \doi{10.1145/3071178.3071205},
  \urlprefix\url{http://doi.acm.org/10.1145/3071178.3071205}

\bibitem[{Rozenberg et~al.(2011)Rozenberg, Bck, and
  Kok}]{rozenberg2011handbook}
Rozenberg G, Bck T, Kok JN (2011) Handbook of natural computing. Springer
  Publishing Company, Incorporated

\bibitem[{Sacks et~al.(1989)Sacks, Welch, Mitchell, and Wynn}]{sacks1989design}
Sacks J, Welch WJ, Mitchell TJ, Wynn HP (1989) Design and analysis of computer
  experiments. Statistical science pp 409--423

\bibitem[{Schwefel(1977)}]{schwefel1977numerische}
Schwefel HP (1977) Numerische Optimierung von Computer-Modellen mittels der
  Evolutionsstrategie: mit einer vergleichenden Einf{\"u}hrung in die
  Hill-Climbing-und Zufallsstrategie. Birkh{\"a}user

\bibitem[{Schwefel(1993)}]{schwefel1993evolution}
Schwefel HPP (1993) Evolution and optimum seeking: the sixth generation. John
  Wiley \& Sons, Inc.

\bibitem[{Shanno(1970)}]{shanno1970conditioning}
Shanno DF (1970) Conditioning of quasi-newton methods for function
  minimization. Mathematics of computation 24(111):647--656

\bibitem[{Shi and Eberhart(1998)}]{shi1998modified}
Shi Y, Eberhart R (1998) A modified particle swarm optimizer. In: Evolutionary
  Computation Proceedings, 1998. IEEE World Congress on Computational
  Intelligence., The 1998 IEEE International Conference on, IEEE, pp 69--73

\bibitem[{Shir and B{\"a}ck(2005)}]{shir2005niching}
Shir OM, B{\"a}ck T (2005) Niching in evolution strategies. In: Proceedings of
  the 7th annual conference on Genetic and evolutionary computation, ACM, pp
  915--916

\bibitem[{Siarry and Berthiau(1997)}]{siarry1997fitting}
Siarry P, Berthiau G (1997) Fitting of tabu search to optimize functions of
  continuous variables. International journal for numerical methods in
  engineering 40(13):2449--2457

\bibitem[{Siarry et~al.(1997)Siarry, Berthiau, Durdin, and
  Haussy}]{siarry1997enhanced}
Siarry P, Berthiau G, Durdin F, Haussy J (1997) Enhanced simulated annealing
  for globally minimizing functions of many-continuous variables. ACM
  Transactions on Mathematical Software (TOMS) 23(2):209--228

\bibitem[{Smit and Eiben(2011)}]{smit2011multi}
Smit S, Eiben A (2011) Multi-problem parameter tuning using bonesa. In:
  Artificial Evolution, pp 222--233

\bibitem[{Snoek et~al.(2012)Snoek, Larochelle, and Adams}]{snoek2012practical}
Snoek J, Larochelle H, Adams RP (2012) Practical bayesian optimization of
  machine learning algorithms. In: Advances in neural information processing
  systems, pp 2951--2959

\bibitem[{Socha and Dorigo(2008)}]{socha2008ant}
Socha K, Dorigo M (2008) Ant colony optimization for continuous domains.
  European journal of operational research 185(3):1155--1173

\bibitem[{S{\o}ndergaard et~al.(2003)S{\o}ndergaard, Madsen, and
  Nielsen}]{sondergaard2003optimization}
S{\o}ndergaard J, Madsen K, Nielsen HB (2003) Optimization using surrogate
  models-by the space mapping technique. PhD thesis, Technical University of
  Denmark

\bibitem[{Stork et~al.(2019)Stork, Zaefferer, and Bartz{-}Beielstein}]{Stor18c}
Stork J, Zaefferer M, Bartz{-}Beielstein T (2019) Improving neuroevolution
  efficiency by surrogate model-based optimization with phenotypic distance
  kernels. In: Kaufmann P, Castillo PA (eds) Applications of Evolutionary
  Computation - 22nd International Conference, EvoApplications 2019, Held as
  Part of EvoStar 2019, Leipzig, Germany, April 24-26, 2019, Proceedings,
  Springer, Lecture Notes in Computer Science, vol 11454, pp 504--519,
  \doi{10.1007/978-3-030-16692-2\_34},
  \urlprefix\url{https://doi.org/10.1007/978-3-030-16692-2\_34}

\bibitem[{Swersky et~al.(2013)Swersky, Snoek, and Adams}]{swersky2013multi}
Swersky K, Snoek J, Adams RP (2013) Multi-task bayesian optimization. In:
  Advances in neural information processing systems, pp 2004--2012

\bibitem[{Talbi(2002)}]{talbi2002taxonomy}
Talbi EG (2002) A taxonomy of hybrid metaheuristics. Journal of heuristics
  8(5):541--564

\bibitem[{Talbi(2009)}]{talbi2009metaheuristics}
Talbi EG (2009) Metaheuristics: from design to implementation, vol~74. John
  Wiley \& Sons

\bibitem[{T{\"o}rn and Zilinskas(1989)}]{Toer89a}
T{\"o}rn A, Zilinskas A (1989) {Global Optimization}. Springer

\bibitem[{Van~Groenigen and Stein(1998)}]{van1998constrained}
Van~Groenigen J, Stein A (1998) Constrained optimization of spatial sampling
  using continuous simulated annealing. Journal of Environmental Quality
  27(5):1078--1086

\bibitem[{Vermetten et~al.(2019)Vermetten, van Rijn, B\"{a}ck, and
  Doerr}]{Vermetten19a}
Vermetten D, van Rijn S, B\"{a}ck T, Doerr C (2019) Online selection of cma-es
  variants. In: Proceedings of the Genetic and Evolutionary Computation
  Conference, ACM, New York, NY, USA, GECCO '19, pp 951--959,
  \doi{10.1145/3321707.3321803},
  \urlprefix\url{http://doi.acm.org/10.1145/3321707.3321803}

\bibitem[{Woeginger(2003)}]{woeginger2003exact}
Woeginger GJ (2003) Exact algorithms for np-hard problems: A survey. In:
  Combinatorial optimization---eureka, you shrink!, Springer, pp 185--207

\bibitem[{Won and Ray(2004)}]{won2004performance}
Won KS, Ray T (2004) Performance of kriging and cokriging based surrogate
  models within the unified framework for surrogate assisted optimization. In:
  Evolutionary Computation, 2004. CEC2004. Congress on, IEEE, vol~2, pp
  1577--1585

\bibitem[{Zlochin et~al.(2004)Zlochin, Birattari, Meuleau, and
  Dorigo}]{zlochin2004model}
Zlochin M, Birattari M, Meuleau N, Dorigo M (2004) Model-based search for
  combinatorial optimization: A critical survey. Annals of Operations Research
  131(1-4):373--395

\end{thebibliography}

\end{document}